\documentclass[sigconf]{acmart}
\usepackage{amsmath}

\usepackage{algorithmic}
\usepackage{graphicx}
\usepackage{textcomp}
\usepackage{xcolor}
\usepackage{url}
\usepackage{booktabs}
\usepackage[linesnumbered,ruled]{algorithm2e}
\usepackage[inkscapeformat=png]{svg}
\usepackage{soul}
\usepackage{enumerate}
\usepackage{enumitem}
\usepackage{xcolor}
\usepackage{multirow}
\usepackage{subfigure}
\usepackage{colortbl}
\usepackage{amsmath}
\usepackage{balance}

\newcommand\Tau{\mathcal{T}}
\newcommand\bl{}

\copyrightyear{2024} 
\acmYear{2024}
\setcopyright{acmlicensed}
\acmConference[KDD '24]{Proceedings of the 30th ACM SIGKDD Conference on Knowledge Discovery and Data Mining}{August 25--29, 2024}{Barcelona, Spain}
\acmBooktitle{Proceedings of the 30th ACM SIGKDD Conference on Knowledge Discovery and Data Mining (KDD '24), August 25--29, 2024, Barcelona, Spain} \acmDOI{10.1145/3637528.3671675}
\acmISBN{979-8-4007-0490-1/24/08}

\begin{document}
\definecolor{lgray}{rgb}{0.9,0.9,0.9}

\title{Dataset Condensation for Time Series Classification via Dual Domain Matching}

\author{Zhanyu Liu}
\affiliation{%
  \institution{Shanghai Jiao Tong University}
  \city{Shanghai}
  \state{China}
}
\email{zhyliu00@sjtu.edu.cn}

\author{Ke Hao}
\affiliation{%
  \institution{Shanghai Jiao Tong University}
  \city{Shanghai}
  \state{China}
}
\email{ke_hao2002@outlook.com}

\author{Guanjie Zheng}
\authornote{Corresponding Author}
\affiliation{%
  \institution{Shanghai Jiao Tong University}
  \city{Shanghai}
  \state{China}
}
\email{gjzheng@sjtu.edu.cn}

\author{Yanwei Yu}
\affiliation{%
  \institution{Ocean University of China}
  \city{Qingdao}
  \country{China}
}
\email{yuyanwei@ouc.edu.cn}

\ccsdesc[500]{Information systems~Data mining}

\keywords{Dataset Condensation; Data Mining; Time Series Classification}
\begin{abstract}
Time series data has been demonstrated to be crucial in various research fields.
The management of large quantities of time series data presents challenges in terms of deep learning tasks, particularly for training a deep neural network.
Recently, a technique named \textit{Dataset Condensation} has emerged as a solution to this problem.
This technique generates a smaller synthetic dataset that has comparable performance to the full real dataset in downstream tasks such as classification.
However, previous methods are primarily designed for image and graph datasets, and directly adapting them to the time series dataset leads to suboptimal performance due to their inability to effectively leverage the rich information inherent in time series data, particularly in the frequency domain.
In this paper, we propose a novel framework named Dataset \textit{\textbf{Cond}}ensation for \textit{\textbf{T}}ime \textit{\textbf{S}}eries \textit{\textbf{C}}lassification via Dual Domain Matching (\textbf{CondTSC}) which focuses on the time series classification dataset condensation task.
Different from previous methods, our proposed framework aims to generate a condensed dataset that matches the surrogate objectives in both the time and frequency domains. 
Specifically, CondTSC incorporates multi-view data augmentation, dual domain training, and dual surrogate objectives to enhance the dataset condensation process in the time and frequency domains.
Through extensive experiments, we demonstrate the effectiveness of our proposed framework, which outperforms other baselines and learns a condensed synthetic dataset that exhibits desirable characteristics such as conforming to the distribution of the original data.
\end{abstract}

\maketitle

\section{Introduction}


The exponential growth of time series data across various domains, such as traffic forecasting~\cite{li2017diffusion,liu2023fdti,liu2023cross,liu2024frequency,liu2024multi,liang2023cblab}, clinical diagnosis~\cite{hayat2022medfuse, Harutyunyan_2019}, and financial analysis~\cite{cheng2022financial,kumbure2022machine}, has presented opportunities for researchers and practitioners. 
However, the sheer volume of data has imposed burdens in terms of storage, processing, and analysis in terms of deep learning context. 
\bl{In the context of some deep learning downstream tasks such as neural architecture search~\cite{rakhshani2020neural} and continual learning~\cite{gupta2021continual}, the utilization of the full dataset for training has the potential to yield bad efficiency. 
Consequently, there is an urgent need to develop effective strategies for condensing time series datasets while preserving their essential information to alleviate the efficiency challenge of training a deep model.}

\begin{figure}[!tp]
    \centering
    \includegraphics[width=\linewidth]{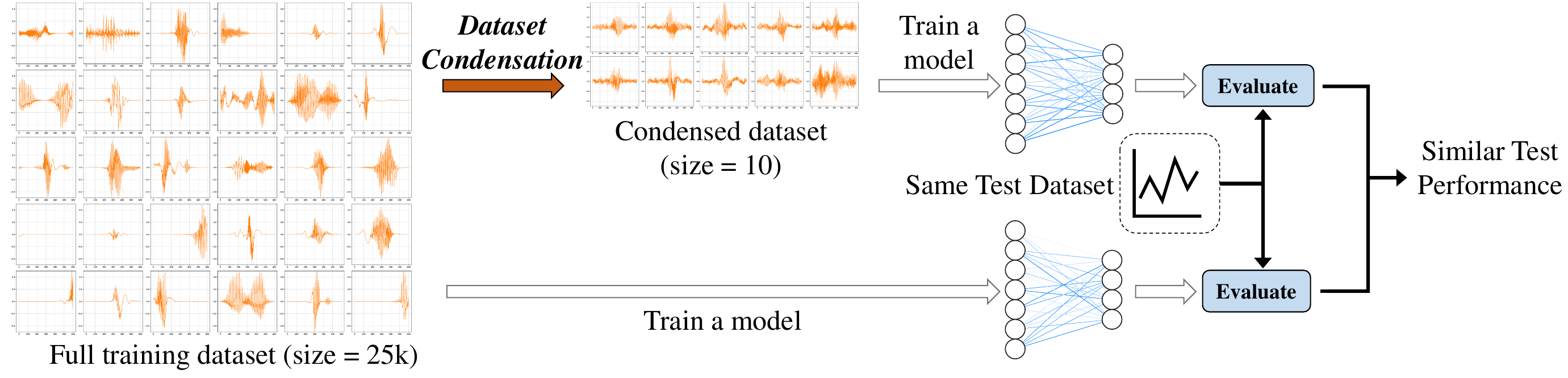}
    \caption{The diagram illustrates the concept of time series data condensation, which aims to learn a small dataset that achieves comparable performance to the full dataset. }
    \label{fig:0_intro}
    \vspace{-.4cm}
\end{figure}

To address this issue, \textit{Dataset Condensation}~\cite{wang2020dataset,zhao2020dataset} has emerged as a powerful and efficient approach.
The core idea is learning a small synthetic dataset that achieves comparable performance to the original full dataset when training the identical model, as depicted in Fig.~\ref{fig:0_intro}.
Unlike traditional data compression techniques, the primary objective of dataset condensation is to distill a reduced-sized synthetic dataset \bl{that could efficiently train a model and effectively generalize to unseen test data} rather than just reduce the size without information loss.
Notably, the fundamental principles of data condensation encompass several strategies aimed at addressing specific surrogate objectives. These strategies include maximizing the testing performance~\cite{nguyen2021dataset,wang2020dataset,nguyen2020dataset,liu2022dataset,deng2022remember,zhou2022dataset,loo2023dataset}, matching the training gradient~\cite{jin2021graph,jin2022condensing,lee2022dataseta,kim2022dataset,zhao2021dataset,zhao2020dataset,du2023minimizing,zhang2023accelerating,cazenavette2022dataset,cui2022scaling,li2022dataset}, and matching the hidden state distribution when training~\cite{lee2022datasetb,wang2022cafe,zhao2023dataset}.

Currently, most of the data condensation research focuses on image~\cite{lee2022dataseta,wang2022cafe,deng2022remember,nguyen2021dataset,zhou2022dataset}, and graph~\cite{jin2022condensing,jin2021graph,liu2022graph,sachdeva2022infinite} \bl{dataset}.
However, the dataset condensation research for time series data is not well explored.
Different from the other modalities, time series data exhibits distinct characteristics, such as periodicity and seasonality, which are closely related to its frequency domain.
The frequency domain plays a crucial role, offering valuable insights and aiding in time series analysis~\cite{woo2022cost,yang2022unsupervised,zhang2022self,zhang2022cross,zhou2022fedformer}.
In the frequency domain, the key patterns and characteristics of time series data that might be difficult to discern in the time domain alone could be identified.
Consequently, there is a pressing need to develop dataset condensation methods specifically tailored for time series data, capitalizing on the rich information available in the frequency domain to enhance the efficiency and effectiveness of condensation techniques.

In this paper, we propose a framework named Dataset \textit{\textbf{Cond}}-ensation for \textit{\textbf{T}}ime \textit{\textbf{S}}eries \textit{\textbf{C}}lassification via Dual Domain Matching (\textbf{CondTSC}).
The proposed framework aims to generate a condensed synthetic dataset by matching the surrogate objectives between synthetic dataset and full real dataset in both the time domain and frequency domain.
Specifically, we first introduce \textbf{Multi-view Data Augmentation} module, which utilizes the data augmentation techniques to project the synthetic data into different frequency-enhanced spaces. 
This approach enriches data samples and strengthens the matching process of surrogate objectives, resulting in synthetic data that is more robust and representative. 
Then, we propose \textbf{Dual Domain Training} module, which leverages both the time domain and frequency domain of the synthetic data. 
By encoding information from both domains, the downstream surrogate matching module benefits from the rich dual-domain information. 
Furthermore, we introduce \textbf{Dual Objectives Matching} module.
By matching the surrogate objectives, training these networks with the condensed synthetic dataset yields similar gradient and hidden state distributions as training with the full real dataset. 
This enables the synthetic data to learn the dynamics of real data when training the networks such as CNN or Transformer.
\bl{Overall, the CondTSC framework generates a condensed synthetic dataset that keeps the dynamics of training a network for time series analysis.} 

Extensive experiments show that CondTSC achieves outstanding performance in the data condensation task in many scenarios such as human activity recognition (HAR) and insect audio classification. For example, we achieve $61.38\%$ accuracy with $0.1\%$ of the original size and $86.64\%$ accuracy with $1\%$ of the original size, compared with $93.14\%$ accuracy with full original size in HAR dataset.

Our contributions can be summarized as follows:
\begin{itemize}
    \item To the best of our knowledge, we are the first to systematically complete the data condensation problem for time series classification. We highlight the importance of the frequency domain in time series analysis and propose to integrate the frequency domain view into the dataset condensation process. This allows for a comprehensive understanding and analysis of time series patterns and behaviors.
    \item We propose a novel framework named CondTSC, which incorporates multi-view data augmentation, dual domain training, and dual surrogate objectives to enhance the dataset condensation process in the time and frequency domains. 
    \item We validate the effectiveness of CondTSC through extensive experiments. \bl{Results show our proposed framework not only outperforms other baselines in the condensed training setting but also shows good characteristics such as conforming to the distribution of the original data}.
\end{itemize}



\section{Related Work}

\begin{figure*}[!t]
    \centering
    \includegraphics[width=0.9\linewidth]{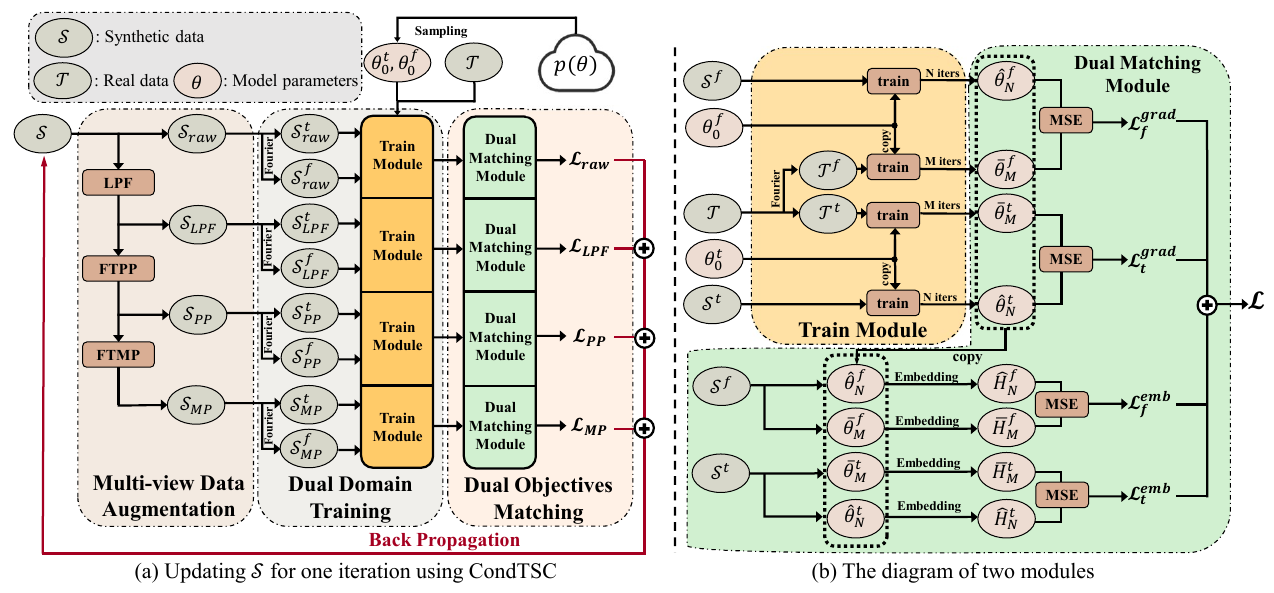}
    \caption{The diagram of CondTSC. LPF indicates low pass filter. FTPP indicates Fourier transform phase perturbation and FTMP indicates Fourier transform magnitude perturbation.}
    \label{fig:main_fig}
    \vspace{-.4cm}
\end{figure*}

\bl{
\subsection{Time Series Compression}
Time series compression aims to compress the time series data without losing the important information, which benefits the process of storage, analysis, and transmission~\cite{chiarot2023time}.
There are many works that focus on this problem by using dictionary compression~\cite{marascu2014tristan}, Huffman coding~\cite{mogahed2018development}, and so on.
However, the compressed dataset does not work efficiently in training a deep network. 
In order to utilize the compressed data, it must first be decoded to its original size, and subsequently, the network iterates through the original-sized dataset to generate loss for back-propagation.
Both of these processes are time-consuming.
Consequently, time series compression is unsuitable for tasks demanding high training efficiency, such as neural architecture search~\cite{JMLR:v20:18-598}.
To handle this problem, a newly emerging area \textit{Dataset Condensation} gives its solution.
}

\subsection{Dataset Condensation}
\bl{\textit{Dataset Condensation} aims to learn a small synthetic dataset that could achieve comparable performance to the original full dataset when training a deep neural network.}
Currently, there are three main approaches to Dataset Condensation as follows.
(1) The testing performance maximization approaches focus on maximizing the performance on the real-world test-split dataset of the network trained by the condensed synthetic dataset~\cite{nguyen2021dataset,wang2020dataset,nguyen2020dataset,liu2022dataset,deng2022remember,zhou2022dataset,loo2023dataset}.
These approaches would face the challenge of the high cost of computing high-order derivatives and thus some of the works utilize the kernel ridge regression~\cite{nguyen2020dataset,nguyen2021dataset} or neural feature regression~\cite{zhou2022dataset} to reduce the computational complexity.
(2) The training gradient matching approach aims to match the training gradient of the model trained on the condensed dataset with that of the model trained on the full dataset~\cite{jin2021graph,jin2022condensing,lee2022dataseta,kim2022dataset,zhao2021dataset,zhao2020dataset,du2023minimizing,zhang2023accelerating,cazenavette2022dataset,cui2022scaling,li2022dataset}. By matching the gradients, the condensed dataset captures the essential full-data training dynamics and enables efficient model training.
Moreover, there are two main surrogate matching objectives including single-step gradient~\cite{jin2021graph,jin2022condensing,lee2022dataseta,kim2022dataset,zhao2021dataset,zhao2020dataset,cui2022scaling,li2022dataset} and multi-step gradients~\cite{du2023minimizing,zhang2023accelerating,cazenavette2022dataset}.
(3) The hidden state distribution matching approach focuses on matching the distribution of hidden states during the training process~\cite{lee2022datasetb,wang2022cafe,zhao2023dataset,sajedi2023datadam}. These approaches have high efficiency due to their low computational complexity.
However, all approaches of these three types focus on images or graphs, lacking detailed analysis for techniques for the time series datasets.

\subsection{Frequency-enhanced Time series analysis}
Recently, several studies have focused on enhancing time series analysis by leveraging the frequency domain.
\bl{MCNN}~\cite{cui2016multi} incorporates multi-frequency augmentation of time series data and utilizes a simple convolutional network.
Frequency-CNN~\cite{an2018automatic} utilizes a simple convolutional network on the frequency domain and gets good results on automatic early detection.
\bl{BTSF}~\cite{yang2022unsupervised} proposes iterative bilinear spectral-time mutual fusion and trains the encoder based on triplet loss.
Recently, many studies have focused on developing frequency-based self-supervised architectures specifically designed for time series analysis. 
CoST~\cite{woo2022cost} and TF-C~\cite{zhang2022self} construct contrastive losses based on the frequency domain pattern to learn more robust representations.
FEDFormer~\cite{zhou2022fedformer} proposes a frequency-based representation learning to improve the effectiveness and efficiency of the Transformer.
The outstanding results achieved by these methods have provided evidence for the effectiveness of incorporating frequency information in time series analysis.

\section{Preliminary}

\subsection{Problem Overview}

\begin{figure*}[!th]
    \centering
    \includegraphics[width=0.83\linewidth]{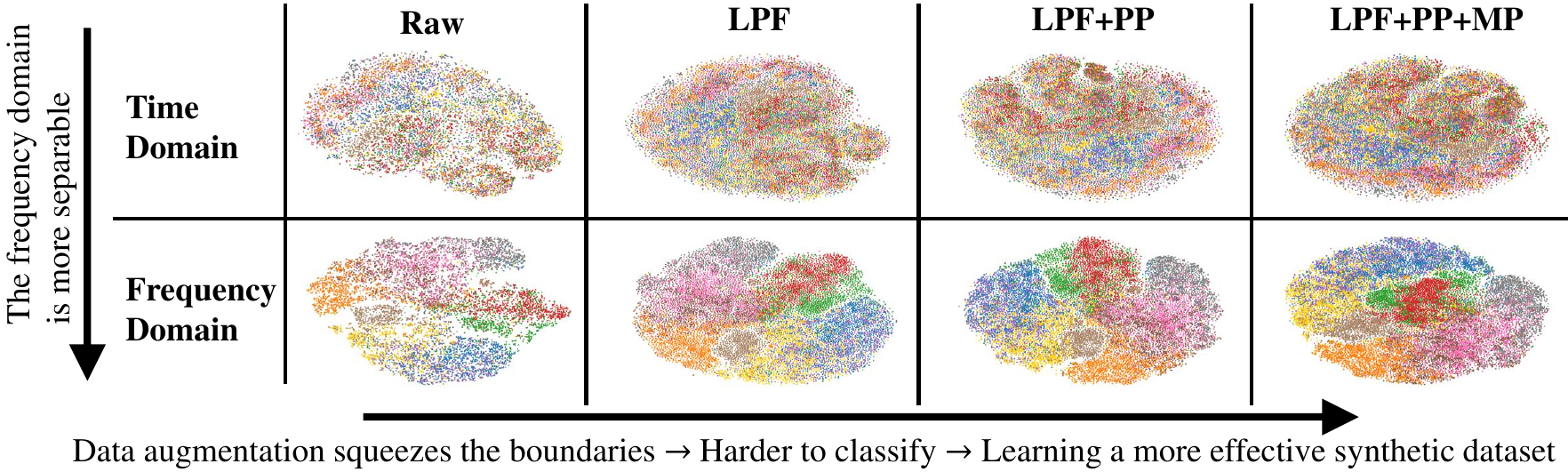}
    \caption{The TSNE visualization of the Insect dataset in both the time domain and the frequency domain. This demonstrates the intuition to do augmentation and utilize the dual-domain information for the time series data. The data in the frequency domain shows better decision boundaries and the data augmentations squeeze the boundaries between different classes.}
    \label{fig:3DualEncoder}
\end{figure*}

The goal of Data Condensation is to learn a synthetic dataset $\mathcal{S}=\{(s_i, y_i)\}_{i=1}^{|\mathcal{S}|}$ from the real training dataset $\mathcal{\Tau}=\{(x_i,y_i)\}_{i=1}^{|\mathcal{\Tau}|}$, where each data $s_i, x_i\in\mathbb{R}^d$, and the label $y_i\in\mathcal{Y}=\{0,1,\ldots,C-1\}$.
We want the size of the synthetic dataset to be much smaller than the size of the real training dataset, i.e., $|\mathcal{S}|\ll|\mathcal{\Tau}|$ while keeping the network trained by synthetic data comparable to the network trained by the real training dataset. 

Formally, by denoting $\widehat{\theta}$ as the network parameter trained by $\mathcal{S}$ and $\bar{\theta}$ as the network parameter trained by $\mathcal{\Tau}$, we could write as follows:
\begin{equation}
    \widehat{\theta} = \mathop{\arg\min}\limits_\theta\sum_{(s_i,y_i)\in\mathcal{S}}\ell(s_i,y_i,f_\theta),
\end{equation}
\begin{equation}
    \bar{\theta} = \mathop{\arg\min}\limits_\theta\sum_{(x_i,y_i)\in\mathcal{\Tau}}\ell(x_i,y_i,f_\theta).
\end{equation}
Then, the aim is to ensure that both networks exhibit comparable generalization capabilities when evaluated on the test dataset $\mathcal{\Tau}_{test}$ as follows:
\begin{equation}
    \mathbb{E}_{(x,y)\in\mathcal{\Tau}_{test}}[Eval(f_{\widehat{\theta}}(x),y)]\simeq\mathbb{E}_{(x,y)\in\mathcal{\Tau}_{test}}[Eval(f_{\bar{\theta}}(x),y)].
\end{equation}
Here, $\ell(\cdot,\cdot,\cdot)$ indicates training loss, the  $f_\theta$ indicates the network parameterized by $\theta$ and $Eval(\cdot,\cdot)$ indicates arbitrary evaluation metrics such as accuracy and recall.

\section{Method}

In this section, we present our proposed framework CondTSC, which comprises three key modules.
The framework's architecture is depicted in Fig.~\ref{fig:main_fig}, and the detailed pseudo code is shown in Alg.~\ref{alg:CondTSC}. 
The first module, termed Multi-view Data Augmentation, projects the synthetic dataset $\mathcal{S}$ into spaces of multi-view to enrich the data samples.
Next, the Dual Domain Encoding module applies the Fourier transform and trains separate networks on each domain to obtain network parameters specific to $\mathcal{S}$ and $\mathcal{\Tau}$.
Lastly, the Dual Objectives Matching module leverages the trained network parameters to construct two types of surrogate objectives: parameter matching and embedding space matching.
By matching these surrogate objectives, the losses computed across all views can be back-propagated to update the synthetic dataset $\mathcal{S}$ accordingly.

\subsection{Initializing $\mathcal{S}$}
In previous research, pixel-level random initialization has been commonly utilized to initialize $\mathcal{S}$, as evidenced by numerous studies~\cite{zhao2020dataset,zhao2021dataset,zhao2023dataset,cazenavette2022dataset,kim2022dataset,jin2021graph,jin2022condensing}. However, these studies have focused primarily on condensing image or graph datasets. In contrast, empirical evidence suggests that random initialization results in poor performance when condensing time-series datasets. Therefore, to address this issue, we propose using K-means to cluster the data samples of each class from the full dataset $\Tau$, and then selecting the clustering centroids of each class as the initial data samples of $\mathcal{S}$.
Formally, it could be formulated as follows.
\begin{equation}
    \mathcal{S}_{i} = Kmeans(\mathcal{\Tau}_{i}, spc)
    \label{eq:initialize}
\end{equation}
In this context, $\mathcal{S}_i$ represents the samples of the $i$-th class within $\mathcal{S}$, while $\mathcal{\Tau}_i$ represents the samples of the $i$-th class within $\mathcal{\Tau}$. The function $Kmeans(X, K)$ employs K-means algorithm on data $X$ with $K$ clustering centroids and outputs the centroids to initialize $\mathcal{S}_i$.

\subsection{Multi-view Data Augmentation}
\noindent\textit{\textbf{Goal:}}
This module essentially projects the synthetic data $\mathcal{S}$ into embedding spaces of multiple views by conducting several data augmentations~\cite{wen2020time} sequentially.
The visualization in Figure~\ref{fig:3DualEncoder} demonstrates the increased separability of the frequency domain, suggesting that augmenting data in the frequency domain yields more effective results.
Consequently, different from~\cite{zhao2021dataset,deng2022remember} that uses augmentations for images such as jittering and cropping, we employ frequency-domain augmentations such as low pass filtering.
The augmentation process, as depicted in Figure~\ref{fig:3DualEncoder}, generates more data and compresses the boundaries between classes in the frequency domain.
As a result, the synthetic data $\mathcal{S}$ can more effectively adhere to the training dynamics such as gradient and embeddings exhibited by the real data $\mathcal{\Tau}$ by virtue of matching the surrogate objectives across different views of the synthetic data $\mathcal{S}$.

\noindent{\textit{\textbf{Augmentations:}}}
As shown in Fig.~\ref{fig:main_fig}(a), we will conduct three frequency-enhanced data augmentations sequentially to $\mathcal{S}$, including Low Pass Filter (LPF), Fourier Transform Phase Perturbation (FTPP), Fourier Transform Magnitude Perturbation (FTMP).

First, by denoting the Fourier Transform and Inverse Fourier Transform as $FT(\cdot)$ and $IFT(\cdot)$ respectively, the first data augmentation LPF could be formally written as follows:
\begin{equation}
    \mathcal{S}_{raw}=\mathcal{S},
    \label{eq:raw}
\end{equation}
\begin{equation}
    \mathcal{S}_{LPF}=IFT(LPF(FT(\mathcal{S}_{raw}))).
    \label{eq:LPF}
\end{equation}
Here $LPF(\cdot)$ indicates a low pass filter.
We directly keep 50\% of the low-frequency components and discard the remaining, thereby ensuring the robustness of the synthetic data while preserving the key information in the frequency domain (above 90\%).
Then, we perturb the phase of the low-passed synthetic data $\mathcal{S}_{LPF}$ as follows:
\begin{equation}
    \mathcal{S}_{PP}=IFT(Noise(FT(\mathcal{S}_{LPF})_{phase})).
    \label{eq:PP}
\end{equation}
Here $FT(\cdot)_{phase}$ indicates the phase of the frequency domain.
To introduce diversity and enhance the robustness of the synthetic data, Gaussian noise is added to the phase components using the $Noise(\cdot)$ function.
This perturbation further expands the variation within the synthetic dataset.
Furthermore, we perturb the magnitude of $\mathcal{S}_{PP}$ as follows:
\begin{equation}
    \mathcal{S}_{MP}=IFT(Noise(FT(\mathcal{S}_{PP})_{magnitude})).
    \label{eq:MP}
\end{equation}
Here $FT(\cdot)_{magnitude}$ indicates the magnitude of the frequency domain.
This perturbation introduces variations in the magnitude of the frequency domain, allowing the synthetic data to capture a wider range of amplitude and intensity characteristics.

In summary, the Multi-view Data Augmentation module in the CondTSC framework plays a crucial role in enriching the data samples and enhancing the matching of surrogate objectives by projecting the synthetic data $\mathcal{S}$ into multiple high-dimensional spaces through sequential data augmentations.

\subsection{Dual Domain Training}

\noindent\textit{\textbf{Goal:}}
As shown in Fig.~\ref{fig:3DualEncoder}, the frequency domain is more separable, which means fusing the frequency domain into the dataset condensation could enhance the performance and effectiveness.
Consequently, this module aims to incorporate both the time and frequency domains and utilize them in the construction of surrogate objectives in the next module.

\noindent\textit{{\textbf{Training Module:}}}
The Training Module receives three types of input: the synthetic data in both the time domain $\mathcal{S}^t$ and the frequency domain $\mathcal{S}^f$, the full real data in both the time domain $\mathcal{\Tau}^t$ and the frequency domain $\mathcal{\Tau}^f$, and the sampled network parameters in both time domain $\theta_0^t$ and the frequency domain $\theta_0^f$ randomly sampled from $p(\theta)$.
\textbf{For simplicity}, we here introduce the training process in the time domain, while the process in the frequency domain is identical.

The training process follows standard training procedures, where we utilize the synthetic data $\mathcal{S}^t$ and the real data $\mathcal{\Tau}^t$ to train the network parameters $\theta_0^t$.
We employ standard training settings, such as stochastic gradient descent or Adam optimization, to update the parameters based on the loss between the network predictions and the ground truth labels.
Furthermore, we train the network with $\mathcal{S}^t$ and $\mathcal{\Tau}^t$ for $N$ and $M$ iterations respectively.
We set $N\ll M$ to avoid overfitting to the small synthetic data $\mathcal{S}^t$.

The output of the Training Module is the trained parameters $\widehat{\theta}^t_N$ (trained by $\mathcal{S}$) and $\bar{\theta}^t_M$ (trained by $\mathcal{\Tau}$). 
These trained parameters play a crucial role in the subsequent modules as they are utilized in the construction of surrogate objectives, which facilitate the matching of training dynamics between the synthetic and real datasets.

\noindent\textit{{\textbf{Model Parameters $p(\theta)$:}}}
As aforementioned, the aim of dataset condensation is to train a network using the synthetic dataset $\mathcal{S}$ that achieves performance comparable to a network trained using the real dataset $\mathcal{\Tau}$.
To achieve this, we will match the surrogate objectives with the given parameter distribution.
However, in practice, computing the exact parameter distribution is infeasible.
Therefore, we collect the parameters from the training trajectories when training a network with $\mathcal{\Tau}$ and utilize them as an approximation for the parameter distribution $p(\theta)$.
Moreover, in line with prior research~\cite{wang2020dataset,zhao2021dataset,zhao2020dataset,cazenavette2022dataset}, we employ a single network structure in the training process, i.e., the used $p(\theta)$ is from a parameter space of single network structure $f$.
Furthermore, we conduct experiments that train $\mathcal{S}$ based on one network structure distribution $p_1(\theta)$ and evaluate the trained $\mathcal{S}$ on another distribution $p_2(\theta)$ to further evaluate the effectiveness of our proposed framework.

\subsection{Dual Domain Surrogate Objective Matching}

\begin{algorithm}[!tp]
    \KwIn{Full real data $\mathcal{\Tau}$, network parameter distribution $p(\theta)$, samples per class $spc$, training epochs $Epo$. 
    }
    \KwOut{Condensed Synthetic data $\mathcal{S}$}
    Initialize $\mathcal{S}_i$ for $i=\{0,1,\ldots,C-1\}$ based on Eq.~\eqref{eq:initialize}\;

    \For{$e\leftarrow range(0,Epo,1)$}{
        $\mathcal{L}\leftarrow0.0,\ \ \  aug\_list\leftarrow[raw, LPF,PP,MP]$\;
        \For{$aug\leftarrow aug\_list$}{
            \tcp{Multi-view Data Augmentation}
            Compute $\mathcal{S}_{aug}$ based on Eqs~\eqref{eq:raw},\eqref{eq:LPF},\eqref{eq:PP},\eqref{eq:MP}\;
            \tcp{Dual Domain Training}
            $\mathcal{S}^t_{aug}\leftarrow\mathcal{S}_{aug},\ \ \ \mathcal{S}^f_{aug}\leftarrow FT(\mathcal{S}^t_{aug})$\;
            $\mathcal{\Tau}^t\leftarrow batch(\mathcal{\Tau}),\ \ \ \mathcal{\Tau}^f\leftarrow FT(\mathcal{\Tau}^t)$\;
            $\theta_0^t,\theta_0^f\leftarrow Sample(p(\theta))$\;
            $\widehat{\theta}_N^t\leftarrow Train\_Niters(\mathcal{S}^t_{aug},\theta_0^t)$\;
            $\widehat{\theta}_N^f\leftarrow Train\_Niters(\mathcal{S}^f_{aug},\theta_0^f)$\;
            $\bar{\theta}_M^t\leftarrow Train\_Miters(\mathcal{\Tau}^t,\theta_0^t)$\;
            $\bar{\theta}_M^f\leftarrow Train\_Miters(\mathcal{\Tau}^f,\theta_0^f)$\;
            \tcp{Dual Objectives Matching}
            \tcp{Surrogate objective 1: multi-step gradient}
            $\mathcal{L}_t^{grad}=\cfrac{||\widehat{\theta}_N^t - \bar{\theta}_M^t||_2^2}{||\theta_0^t-\bar{\theta}_M^t||_2^2},\ \ \ \mathcal{L}_f^{grad}=\cfrac{||\widehat{\theta}_N^f - \bar{\theta}_M^f||^2_2}{||\theta_0^f-\bar{\theta}_M^f||_2^2}$\;
            \tcp{Surrogate objective 2: embedding space}
            Compute $\mathcal{L}_t^{emb}, \mathcal{L}_f^{emb}$ based on Eqs~\ref{eq:embt},\ref{eq:embf}\;
            $\mathcal{L}=\mathcal{L} + \mathcal{L}_t^{grad}+\mathcal{L}_f^{grad}+\lambda(\mathcal{L}_t^{emb}+\mathcal{L}_f^{emb})$
        }
        Back-propagate $\mathcal{L}$ and update $\mathcal{S}$
    }

    \Return $\mathcal{S}$
    
\caption{Pseudo code for CondTSC.}
\label{alg:CondTSC}
\end{algorithm}

\noindent\textit{\textbf{Goal:}}
This module aims to generate surrogate objectives from different views using both the synthetic dataset $\mathcal{S}$ and the real dataset $\mathcal{\Tau}$, and subsequently match these objectives.
By computing the matching loss across the different views, we can update the condensed dataset $\mathcal{S}$.
After the surrogate objectives are well matched, the synthetic dataset $\mathcal{S}$ could be considered as the condensed dataset of the original dataset $\mathcal{\Tau}$ for learning the downstream task.

\noindent\textit{\textbf{Gradient Matching:}}
The aim of gradient matching is to make the gradient of $\mathcal{S}$ and $\mathcal{\Tau}$ similar.
By doing so, training a network with $\mathcal{S}$ could generate a similar trained model, and thus the performance could be comparable.
Here we take the multi-step gradient of training a network parameterized by $\theta_0$ as the surrogate objective similar to~\cite{cazenavette2022dataset,du2023minimizing,zhang2023accelerating}.
Intuitively, we directly minimize the mean square error between these gradients.
Taking the parameter in the time domain as an example, the error could be formulated as follows:
\begin{equation}
\begin{aligned}
    \mathcal{L}_t^{grad} &=  \cfrac{||grad_{\mathcal{S}}-grad_{\mathcal{\Tau}}||_2^2}{||grad_\mathcal{\Tau}||_2^2}.
\end{aligned}
\end{equation}
Here, we use the final network parameter $\widehat{\theta}_N^t$ to minus the initial network parameter $\theta_0^t$ to get the gradient $grad_{\mathcal{S}}=\widehat{\theta}_N^t-\theta_0^t$. 
Moreover, $grad_{\mathcal{\Tau}}$ could be similarly calculated.
Consequently, we get the gradient matching loss in the time domain as follows:
\begin{equation}
\begin{gathered}
\begin{aligned}
  \mathcal{L}_t^{grad}  & =   \cfrac{||(\widehat{\theta}_N^t-\theta_0^t) - (\bar{\theta}_M^t-\theta_0^t)||_2^2}{||\bar{\theta}_M^t-\theta_0^t||_2^2}\\
   & =  \cfrac{||\widehat{\theta}_N^t-\bar{\theta}_M^t||_2^2}{||\theta_0^t-\bar{\theta}_M^t||_2^2}.
\end{aligned}
\end{gathered}
\end{equation}
Furthermore, the gradient matching loss in the frequency domain could be similarly formulated as follows
\begin{equation}
\begin{gathered}
\begin{aligned}
    \mathcal{L}_f^{grad}  & =\cfrac{||\widehat{\theta}_N^f - \bar{\theta}_M^f||^2_2}{||\theta_0^f-\bar{\theta}_M^f||_2^2}.
\end{aligned}
\end{gathered}
\end{equation}

\noindent\textit{\textbf{Parameter Space Matching}}
Previous methods such as those described in~\cite{lee2022datasetb,wang2022cafe,zhao2023dataset} aimed to match the distribution of embedding between $\mathcal{S}$ and $\mathcal{\Tau}$ using maximum mean discrepancy (MMD)\cite{zhao2023dataset,tolstikhin2016minimax,gretton2012kernel}.
However, these methods use random model parameters $\theta_{rand}$ and thus did not utilize the rich information of training dynamics contained in the space of trained parameters $\widehat{\theta}_N^t$ and $\bar{\theta}_M^t$.
By matching these spaces, these parameters could generate similar embeddings, and thus the training direction of $\widehat{\theta}_N$ could be better guided.
To match these spaces, we first compute the embedding of $\mathcal{S}$ in these spaces
\begin{equation}
    \widehat{H}_N^t=f_{\widehat{\theta}^t_N}(\{s_i|(s_i,y_i)\in\mathcal{S}^t\}),\bar{H}_M^t=f_{\bar{\theta}^t_M}(\{s_i|(s_i,y_i)\in\mathcal{S}^t\}),
\end{equation}
\begin{equation}
    \widehat{H}_N^f=f_{\widehat{\theta}^f_N}(\{s_i|(s_i,y_i)\in\mathcal{S}^f\}),\bar{H}_M^f=f_{\bar{\theta}^f_M}(\{s_i|(s_i,y_i)\in\mathcal{S}^f\}).
\end{equation}

Then, we match the position-wise mean of the embedding of the time domain parameters and the frequency domain parameters separately, i.e., the mean that converts tensor size from $(B, D)$ to $(D)$ where $B$ is the batch size and $D$ is the embedding size as follows:
\begin{equation}
    \mathcal{L}_t^{emb}=||Mean(\widehat{H}_N^t) - Mean(\bar{H}^t_M)||_2^2,
    \label{eq:embt}
\end{equation}
\begin{equation}
    \mathcal{L}_f^{emb}=||Mean(\widehat{H}_N^f) - Mean(\bar{H}^f_M)||_2^2.
    \label{eq:embf}
\end{equation}
Finally, the total loss could be summarized as follows:
\begin{equation}
    \mathcal{L}_{aug}=\mathcal{L}_t^{grad}+\mathcal{L}_f^{grad}+\lambda(\mathcal{L}_t^{emb}+\mathcal{L}_f^{emb}).
\end{equation}
Here $aug\in\{raw,LPF,PP,MP\}$ indicates different views of synthetic data as shown in Fig.~\ref{fig:main_fig} and $\lambda$ is a scaling factor between the gradient loss and the embedding space matching loss.
We then summarize the losses of all views as follows.
\begin{equation}
    \mathcal{L} = \mathcal{L}_{raw}+\mathcal{L}_{LPF}+\mathcal{L}_{PP}+\mathcal{L}_{MP}
\end{equation}
The loss could be summed and back-propagated to $\mathcal{S}$ to make it better capture the training dynamics of $\mathcal{\Tau}$.

\section{Experiment}

\begin{table}[!t]
    \caption{Details of time series datasets.}
    \label{tab:dataset}
    \centering
    \resizebox{0.76\linewidth}{!} {
    \begin{tabular}{c|c|c|c|c|c}
    \toprule
    Dataset & \# Train & \# Test &  Length & \# Channel & \# Class\\
    \midrule
    HAR & 7,352 & 2,947 & 128 & 9 & 6 \\
    Electric & 8,926 & 7,711 & 96 & 1&  7\\
    Insect & 25,000 & 25,000 & 600 & 1  & 10\\
    FD & 8,184 & 2,728 & 5,120 & 1 & 3 \\
    Sleep & 25,612 & 8,910 & 3,000 & 1 & 5 \\
    \bottomrule
    \end{tabular}
    }
    \vspace{-.4cm}
\end{table}

\begin{table*}[!t]
\caption{Overall \bl{accuracy(\%)} performance. The experiment is conducted with the CNN network with batch normalization. The best result is highlighted in bold and grey in each row, and the second-best result is underlined. Marker * indicates the mean of the results is statistically significant  (* means t-test with p-value $<$ 0.01).}
\centering
\label{tab:overall}
\resizebox{1\linewidth}{!}{
\begin{tabular}{c c c|c c c| c c c c c c c | c| c }
\toprule
& $spc$ & ratio (\%) & Random & K-means & Herding & DD & DC & DSA & DM &MTT & IDC & HaBa & CondTSC & Full (upper bound)\\
\midrule
\midrule
\multirow{3}{*}{HAR} & 1 & 0.1  & 44.38$\pm$4.27 & 50.97$\pm$4.30 & 56.09$\pm$2.87 & 21.29$\pm$6.32 & \underline{58.30$\pm$3.32} & 55.41$\pm$6.67 & 52.87$\pm$4.64 & 50.36$\pm$3.85 & 40.93$\pm$4.62  & 45.53$\pm$6.44 & \cellcolor{lgray}{\textbf{61.38$\pm$3.71*}} & \multirow{3}{*}{93.14$\pm$1.03} \\
& 5 & 0.5  &  62.98$\pm$2.21  & 72.19$\pm$1.83 & 64.37$\pm$1.85 & 18.71$\pm$5.27 & 65.83$\pm$3.97 & 65.16$\pm$3.67 & 70.82$\pm$4.63 & \underline{77.66$\pm$3.86} & 54.41$\pm$5.02
  & 56.22$\pm$5.56 & \cellcolor{lgray}{\textbf{82.20$\pm$2.68*}} & \\
& 10 & 1  & 68.77$\pm$1.82 & 76.31$\pm$2.24& 69.37$\pm$2.49 & 19.64$\pm$6.78 & 67.50$\pm$5.15 & 70.99$\pm$3.46 & 78.25$\pm$3.43 & \underline{83.66$\pm$2.65} & 59.99$\pm$3.37
 &  60.37$\pm$6.43 & \cellcolor{lgray}{\textbf{86.64$\pm$2.10*}} & \\
\midrule

\multirow{3}{*}{Electric} & 1 & 0.1   &38.40$\pm$4.28  & 39.07$\pm$2.67 &32.10$\pm$3.61 & 34.95$\pm$6.38  & 47.13$\pm$5.74 & \underline{47.63$\pm$6.05} & 44.86$\pm$6.05 & 39.03$\pm$3.20 &  42.56$\pm$3.60
 & 41.91$\pm$3.24 & \cellcolor{lgray}{\textbf{48.84$\pm$2.18*}} & \multirow{3}{*}{68.35$\pm$0.79}\\

& 5 & 0.5   & 45.55$\pm$3.91 & 46.87$\pm$2.62 &35.46$\pm$4.65 & 35.09$\pm$8.16& 50.52$\pm$3.58 & \underline{51.76$\pm$2.67} &51.22$\pm$2.85 & 49.09$\pm$3.99 & 49.52$\pm$5.82  &44.32$\pm$2.53 & \cellcolor{lgray}{\textbf{56.17$\pm$1.91*}} & \\

& 10 &  1  & 47.83$\pm$1.88 & 48.76$\pm$2.10 &47.92$\pm$2.48 & 35.69$\pm$8.23& 50.37$\pm$2.66 & 52.06$\pm$4.15 & \underline{52.75$\pm$3.74} & 51.29$\pm$2.97 & 50.34$\pm$5.08 &46.71$\pm$2.23 & \cellcolor{lgray}{\textbf{57.86$\pm$1.89*}} & \\
\midrule

\multirow{3}{*}{Insect}& 1 & 0.05   & 14.26$\pm$0.65  & \underline{16.70$\pm$0.95} &10.42$\pm$0.57 & 10.61$\pm$0.74 & 15.48$\pm$2.09 & 15.12$\pm$1.89 & 12.24$\pm$0.86 & 15.54$\pm$1.75& 13.63$\pm$0.61
 &14.63$\pm$0.72  & \cellcolor{lgray}{\textbf{45.15$\pm$2.99*}} & \multirow{3}{*}{70.78$\pm$1.19} \\
&  10 &  0.5   & 23.34$\pm$1.16  & 25.55$\pm$1.07 & 19.40$\pm$0.59 &10.15$\pm$0.41  & 21.56$\pm$2.12& 21.63$\pm$2.13& 20.60$\pm$1.02  & \underline{28.17$\pm$4.65} & 21.29$\pm$1.32 & 22.83$\pm$1.50 &\cellcolor{lgray}{\textbf{63.75$\pm$0.58*}} &\\
&  20 & 1    & 30.29$\pm$1.28 & 30.10$\pm$1.22 & 23.88$\pm$0.78& 10.18$\pm$0.30 & 24.63$\pm$4.11 & 24.12$\pm$5.50& 25.11$\pm$1.24  & \underline{33.83$\pm$3.04}&  25.12$\pm$1.96 & 26.62$\pm$2.48 &\cellcolor{lgray}{\textbf{64.69$\pm$0.60*}} &\\
\midrule

\multirow{3}{*}{FD}& 1 & 0.036   & 46.24$\pm$2.89  &36.48$\pm$9.52 &43.79$\pm$6.60 & 33.93$\pm$16.4 & 32.03$\pm$16.0  & 35.50$\pm$17.8  & 47.19$\pm$5.73 & 44.41$\pm$6.81 & 45.91$\pm$1.67  & \underline{49.87$\pm$2.15} & \cellcolor{lgray}{\textbf{71.00$\pm$7.12*}} &\multirow{3}{*}{98.51$\pm$0.35}  \\
& 10 &  0.36  &  54.99$\pm$1.96 & 54.06$\pm$3.80 &58.58$\pm$2.07 & 29.00$\pm$18.3&44.02$\pm$16.8 & 41.15$\pm$12.9 & 59.62$\pm$4.01 & \underline{68.29$\pm$4.00}& 53.75$\pm$5.96  & 51.43$\pm$1.03 & \cellcolor{lgray}{\textbf{83.25$\pm$1.76*}}&\\
& 20 &  0.72  & 60.72$\pm$2.37 & 57.74$\pm$4.19 &59.68$\pm$1.76 & 33.47$\pm$16.3 & 45.73$\pm$15.5 &45.46$\pm$0.01 & 64.75$\pm$3.64 & \underline{74.63$\pm$3.95}& 61.77$\pm$7.75 & 50.18$\pm$2.03  & \cellcolor{lgray}{\textbf{90.78$\pm$1.36*}}  & \\
\midrule

\multirow{3}{*}{Sleep} & 1 & 0.02   &  22.68$\pm$5.01 & 31.62$\pm$7.09 &22.46$\pm$6.52 & 23.43$\pm$6.94&22.18$\pm$6.73 &22.98$\pm$6.98 &30.03$\pm$8.88 & 30.39$\pm$4.96&  28.24$\pm$4.97
 &  \underline{31.72$\pm$3.98}  & \cellcolor{lgray}{\textbf{49.71$\pm$7.53*}}& \multirow{3}{*}{80.01$\pm$1.12}\\
& 10 & 0.2   & 40.35$\pm$3.90  & 36.88$\pm$3.80 & 27.09$\pm$3.71& 22.53$\pm$5.97 & 23.90$\pm$3.97 &23.82$\pm$3.93 & 36.08$\pm$5.92 & \underline{47.61$\pm$5.81}& 30.03$\pm$5.50  & 40.13$\pm$3.34 &\cellcolor{lgray}{\textbf{57.63$\pm$2.25*}}& \\
& 50 &  1  & 57.10$\pm$4.84  & 45.88$\pm$2.93 & 30.69$\pm$3.75 &22.82$\pm$5.65 & 24.31$\pm$6.42 &25.57$\pm$4.53 & 40.09$\pm$4.51 & \underline{60.52$\pm$2.41}& 34.59$\pm$5.60 &  55.12$\pm$5.67  &  \cellcolor{lgray}{\textbf{66.46$\pm$1.00*}}  & \\


\bottomrule
\end{tabular}
}
\end{table*}



\subsection{Datasets}
In line with previous research~\cite{wang2020dataset,wang2022cafe,zhao2020dataset}, which utilizes small benchmark datasets for image classification dataset condensation task, we include five public real-world benchmark datasets as shown in Table~\ref{tab:dataset}.
(1) {\textbf{HAR~\cite{anguita2013public}:}} The Human Activity Recognition (HAR) dataset comprises recordings of 30 individuals who volunteered for a health study and engaged in six different daily activities. 
(2) {\textbf{Electric}~\cite{dau2019ucr}:} The ElectricDevice dataset comprises measurements obtained from 251 households, with a sampling frequency of two minutes. 
(3) {\textbf{Insect~\cite{chen2014flying}:}} The InsectSound dataset comprises 50,000 time series instances, with each instance generated by a single species of fly. 
(4) {\textbf{Sleep~\cite{goldberger2000physiobank}:}} The Sleep Stage Classification dataset comprises recordings of 20 people throughout a whole night with a sampling rate of 100 Hz. 
(5) {\textbf{FD~\cite{lessmeier2016condition}:}} The Fault Diagnosis (FD) dataset contains sensor data from a bearing machine under four different conditions, and we chose the first one.

\subsection{Experiment Setting}

\noindent{\textbf{Evaluation protocol:}}
We follow the evaluation protocol of previous studies~\cite{zhao2020dataset,zhao2021dataset,zhao2023dataset}. 
Concretely, it contains three stages sequentially: 
(1) train a synthetic data $\mathcal{S}$. 
(2) train a randomly initialized network of the same structure on the trained synthetic data $\mathcal{S}$ using the same optimization setting ( with 1e-3 learning rate and 300 training epochs). 
(3) Evaluate the trained network on the same test data with the same evaluation setting. 
In all experiments, this process would be repeated five times to reduce the randomness of the results. 
We report the mean and the standard deviation of the results.
In line with prior research~\cite{nguyen2021dataset,wang2020dataset,liu2022dataset,deng2022remember,zhou2022dataset,loo2023dataset,jin2021graph,jin2022condensing,kim2022dataset,zhao2021dataset,zhao2020dataset,zhao2023dataset} that utilize simple deep models to verify the effectiveness of the proposed method, we train the synthetic data based on a wide range of basic models.
The details of the network architecture and the parameter searching are in the Appendix~\ref{sec:exp_details}.
In summary, we believe the comparison between our proposed method and baseline methods is fair.

\noindent{\textbf{Hyper-parameters:}}
There are only a small number of hyper-parameters associated with CondTSC.
For simplicity, we use the same set of hyper-parameters for all datasets.
For the training process in Train Module, we use a learnable learning rate of 1e-3 following previous settings~\cite{cazenavette2022dataset,zhang2023accelerating,du2023minimizing}.
Moreover, we set $N=10$ and $M=1000$ for the number of the training iterations for $\mathcal{S}$ and $\mathcal{\Tau}$ respectively.
For the loss in Dual Matching Module, we set $\lambda=1.0$.
For the update of the synthetic data $\mathcal{S}$, we use an SGD optimizer with a learning rate of $lr_{\mathcal{S}}=1.0$ and update epochs $Epo=2000$.
The whole experiment is implemented by Pytorch 1.10.0 on RTX3090.
The code and data are available in~\url{https://github.com/zhyliu00/TimeSeriesCond}.

\subsection{Overall Performance}

\noindent\textbf{Baselines:}
We select ten baselines to evaluate the performance of CondTSC on the time series data condensation task.
These baselines comprise both coreset selection methods and previous SOTA data condensation methods.
\begin{itemize}[leftmargin=*]
    \item Coreset selection methods: We randomly select data samples (\textbf{Random}), choose the K-means clustering centroids (\textbf{K-means}), add samples of each class to the coreset greedily (\textbf{Herding}).
        
    \item Data condensation methods:
    Since there is no method designed for time series data condensation, we select several data condensation methods for image data.
    \textbf{DD~\cite{wang2020dataset}} maximizes the performance of the network trained by the synthetic data. \textbf{DC~\cite{zhao2020dataset}} matches the one-step training gradient of the synthetic data and the real data. \textbf{DSA~\cite{zhao2021dataset}} implements siamese augmentation based on DC. \textbf{DM~\cite{zhao2023dataset}} matches the embedding distribution of the synthetic data and the real data. \textbf{MTT~\cite{cazenavette2022dataset}} matches the multi-step training gradient. \textbf{IDC~\cite{kim2022dataset}} uses efficient parameterization to improve the space utilization. \textbf{HaBa~\cite{liu2022dataset}} decomposes the synthetic data into bases and hallucinators.
\end{itemize}

\begin{table*}[!th]
    \centering
    \caption{\bl{The Neural Architecture Search (NAS) result. We implement this experiment on the HAR dataset with the search space of 324 ConvNets. The condensed dataset has a size of 60 (10 samples/class, 1\% condensed ratio).}}
    \label{tab:NAS}
    \resizebox{0.91\textwidth}{!}{
    \begin{tabular}{c|c c c c c c c c c c|c}
    \toprule
     &Random & K-means & Herding  & DC & DSA & DM &MTT & IDC & HaBa & CondTSC & Full \\
    \midrule
    \midrule
    Accuracy& 87.70$\pm$0.80 &89.59$\pm$1.09 &89.91$\pm$0.75 &90.38$\pm$0.62 &90.68$\pm$1.05 &90.04$\pm$1.38 &90.07$\pm$0.97 &91.01$\pm$0.66 &82.99$\pm$0.59 &\cellcolor{lgray}{\textbf{92.71$\pm$0.81}} &93.74$\pm$0.78\\
    Correlation& 0.587 &0.547 &0.482 &0.560 &0.307 &0.528 &0.582 &0.364 &-0.413 &\cellcolor{lgray}{\textbf{0.665}} &1.000  \\
    Time cost(min) & 24.49 & 24.69 & 24.53 & 24.45 & 24.49 & 24.45 & 24.50 & 24.39 &144.89 & 24.36  &667.91 \\
    Training samples& 60& 60& 60& 60& 60& 60& 60& 60& 60& 60& 7,352\\
    
    \bottomrule
    \end{tabular}
    }
    
\end{table*}

\noindent\textbf{Overall Performance}
Table~\ref{tab:overall} shows the accuracy of baselines and CondTSC.
Ratio (\%) indicates the condensed ratio which is $size(\mathcal{S})/size(\mathcal{\Tau})$ and $spc$ indicates the number of samples per class of the synthetic data. 
Full indicates the performance of the network trained by full data.
We report the mean and the standard deviation of the accuracy of five experiments with different random seeds. 

From Table~\ref{tab:overall}, the following observations could be drawn:
(1) CondTSC achieves superior performance compared to the baselines.
The accuracy gap between the condensed data and full data is comparable to the data condensation task in computer vision~\cite{zhao2021dataset,zhao2020dataset,zhao2023dataset}, and the condensed data is enough for downstream tasks such as Neural Architecture Search, which we will introduce later.
(2) The result of CondTSC is significant in all settings, and CondTSC could approach the performance of full data by using only $1\%$ of the size of the real data.
(3) Coreset selection methods achieve stable but mediocre performance.
This indicates that directly selecting some samples of the real data gives a base but not global information about the real data.
(4) The performance of the data condensation methods is not as satisfactory as the performance of the image dataset condensation task. 
This highlights the significant disparity between the task of condensing time series data and image data.

\begin{figure}[!tp]
    \centering
    \includegraphics[width=0.91\linewidth]{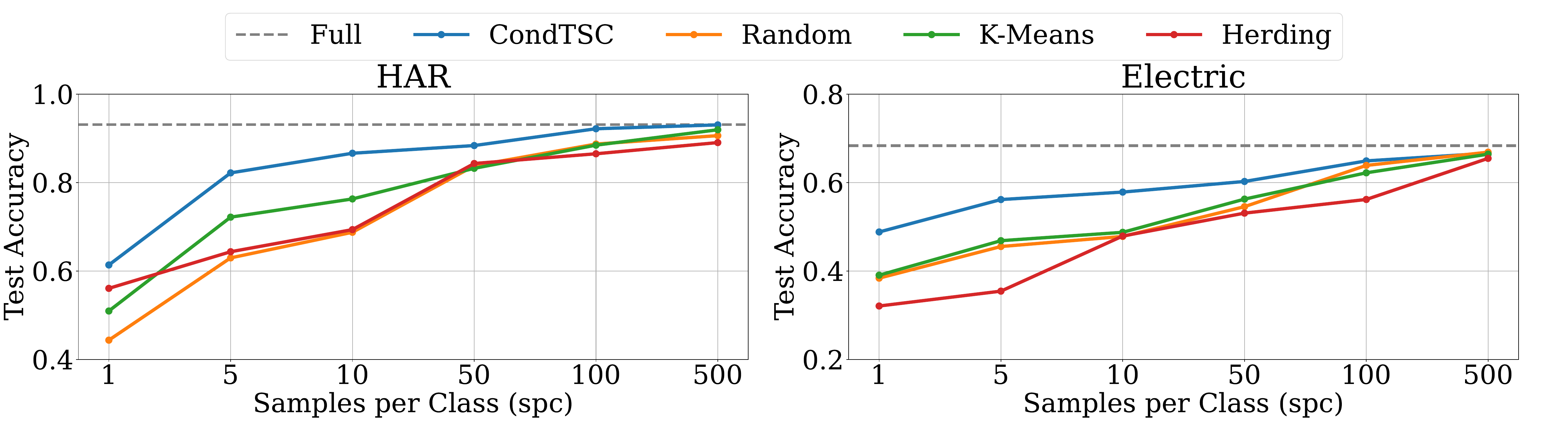}
    \caption{\bl{The accuracy(\%) with larger condensed data size.}}
    \label{fig:5_largeipc}
    \vspace{-.4cm}
\end{figure}

\bl{Furthermore, we evaluate the performance of CondTSC and coreset selection methods with larger condensed data sizes as shown in Fig.~\ref{fig:5_largeipc}. We can observe that increasing the data size makes the methods approach the upper bound and CondTSC has the highest performance. The users could make a trade-off between the accuracy and the training cost according to their downstream tasks.}

\begin{figure*}[!tp]
    \centering
    \includegraphics[width=0.89\linewidth]{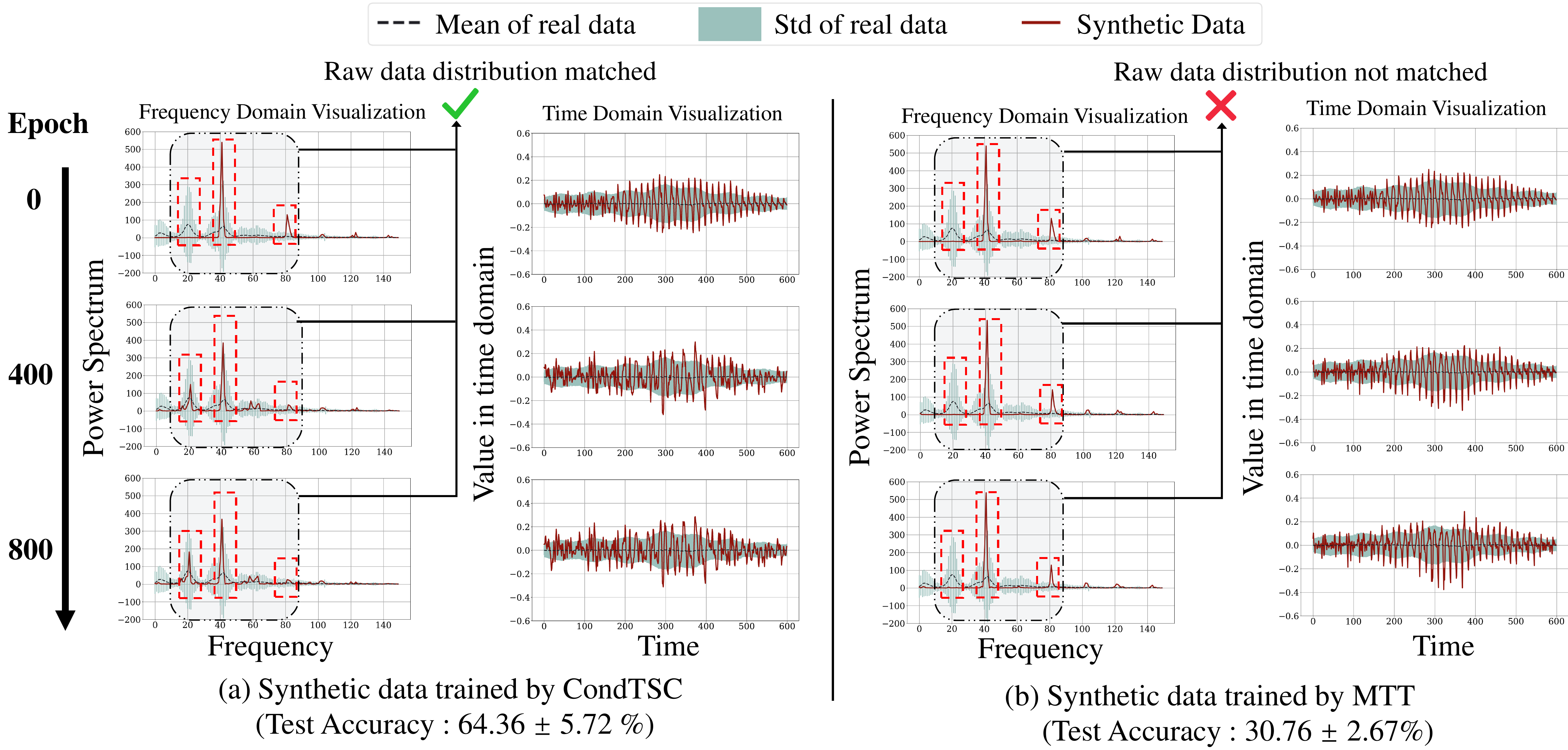}
    \caption{the frequency domain and time domain visualization on Insect dataset of the learning process of synthetic data trained by CondTSC and MTT separately.
    We could observe that the synthetic data trained by CondTSC conforms to the distribution of the real data and consequently achieves remarkable performance.}
    \label{fig:1visualize}
\end{figure*}

\subsection{Downstream Task: NAS}

\bl{
The condensed dataset could be used as a proxy dataset for Neural Architecture Search (NAS)~\cite{elsken2019neural}.
Following the setting of previous research~\cite{zhao2023dataset,zhao2021dataset}, we implement NAS to search the best architecture of 324 Convolutional neural networks on the HAR dataset.
The space is Depth $\in\{2, 3, 4\}$, Width $\in$\{ 32, 64, 128 \}, Normalization $\in$\{ None, BatchNorm, InstanceNorm, LayerNorm \}, Activation $\in$\{ Sigmoid, ReLU, LeakyReLU \}, Pooling $\in$\{ None, Max, Mean \}.
We use 10 samples/class condensed datasets generated by each method.
}

\bl{
The result is shown in Table~\ref{tab:NAS}.
We report 1) the test performance of the model of the best-selected architecture. 2) Spearman's rank correlation coefficient between the validation accuracy obtained from the condensed dataset and the full dataset. We select the top 100 accuracies to filter the outliers following previous research~\cite{zhao2023dataset,zhao2021dataset}. 3) The training time on an RTX3090 GPU and the number of training samples.
We can observe that CondTSC finds the best architecture (92.71\%) and performance correlation to the full dataset (0.665) while decreasing the time cost (from 667 to 24 minutes) and training samples (7352 samples to 60 samples).
The condensing cost is one time off and negligible when training thousands of possible architectures in NAS.
This result indicates that the condensed data could efficiently and effectively accelerate the process of NAS and get good architecture.
Besides, the performance and efficiency of HaBa are bad due to its factorization-based method.
}

\subsection{Ablation Study}
\begin{table}[!t]
\centering
\caption{Ablation Study. In each row, the best \bl{accuracy(\%)} result is highlighted in bold and grey.}
\label{tab:ablation}
\resizebox{0.88\linewidth}{!}{
\begin{tabular}{c c c|c c c c}
\toprule
& ratio(\%) & I/C &Base &Base+A & Base+A+T &Base+A+T+M\\
\midrule
\midrule
\multirow{3}{*}{HAR} & 0.1 & 1 &50.36$\pm$3.85 & 55.63$\pm$2.23 & 60.61$\pm$3.49 & \cellcolor{lgray}{\textbf{61.38$\pm$3.71}} \\
& 0.5 & 5 &77.66$\pm$3.86 &80.38$\pm$1.85 & 80.93$\pm$3.17 & \cellcolor{lgray}{\textbf{82.20$\pm$2.68}}  \\
& 1 & 10 &83.66$\pm$2.65 &84.71$\pm$2.61&85.43$\pm$2.05 & \cellcolor{lgray}{\textbf{86.64$\pm$2.11}} \\
\midrule

\multirow{3}{*}{Electric}&  0.1 & 1 & 39.03$\pm$3.20  & 42.12$\pm$4.11 &  47.53$\pm$3.21  & \cellcolor{lgray}{\textbf{48.84$\pm$2.18}} \\

&0.5& 5  & 49.09$\pm$3.99 & 50.14$\pm$2.30  & 55.67$\pm$1.29 &  \cellcolor{lgray}{\textbf{56.17$\pm$1.91}}  \\

&1& 10  & 51.29$\pm$2.97 &  55.59$\pm$1.72 & 57.20$\pm$1.21 & \cellcolor{lgray}{\textbf{57.86$\pm$1.89}}   \\
\midrule

\multirow{3}{*}{Insect} & 0.05  & 1 & 15.54$\pm$1.75  & 18.58$\pm$1.72& 44.64$\pm$2.97 & \cellcolor{lgray}{\textbf{45.15$\pm$2.99}}  \\
&0.5& 10  & 28.17$\pm$4.65 & 39.85$\pm$2.19   & 62.83$\pm$0.58 &  \cellcolor{lgray}{\textbf{63.75$\pm$0.58}}   \\
&1& 20  & 33.83$\pm$3.04 & 46.90$\pm$1.79  & 63.93$\pm$0.55 & \cellcolor{lgray}{\textbf{64.69$\pm$0.60}} \\
\midrule

\multirow{3}{*}{Sleep} &  0.02 & 1 & 30.39$\pm$4.96  & 33.31$\pm$3.12 & 46.09$\pm$7.64 & \cellcolor{lgray}{\textbf{49.71$\pm$7.53}}  \\
&  0.2 & 10 & 47.61$\pm$5.81  & 49.20$\pm$4.18 & 55.26$\pm$3.27 &  \cellcolor{lgray}{\textbf{57.63$\pm$2.25}}   \\
&  1  & 50 & 60.52$\pm$2.41  & 63.43$\pm$1.11 & 65.86$\pm$1.22 &  \cellcolor{lgray}{\textbf{66.46$\pm$1.00}}   \\
\midrule

\multirow{3}{*}{FD} &  0.036 & 1 & 44.41$\pm$6.81  & 51.41$\pm$3.72 & 69.62$\pm$5.23 & \cellcolor{lgray}{\textbf{71.00$\pm$7.12}}  \\
&0.36& 10  & 68.29$\pm$4.00 & 72.48$\pm$4.13 & 81.11$\pm$2.66  & \cellcolor{lgray}{\textbf{83.25$\pm$1.76}}    \\
&0.72& 20  & 74.63$\pm$3.95 & 77.38$\pm$3.38 & 86.44$\pm$2.15 & \cellcolor{lgray}{\textbf{90.78$\pm$1.36}}     \\
\bottomrule
\end{tabular}
}
\vspace{-.3cm}
\end{table}

In this part, we evaluate the performance of each module of CondTSC to verify the effectiveness of the proposed module.
We denote the Base framework as the simple multi-step gradient matching framework~\cite{cazenavette2022dataset} that directly matches the multi-step training gradient of the synthetic data $\mathcal{S}$ and the real data $\mathcal{\Tau}$.
Then we add the Multi-view Data Augmentation module (A), Dual Domain Training module (T), and Dual Objectives Matching module(M) to the Base framework sequentially and report their mean and standard deviation of accuracy on five experiments with random seeds.

The results are shown in Table~\ref{tab:ablation}.
We can observe that by sequentially adding the modules, the performance is getting better.
By adding the Multi-view Data Augmentation module (Base+A), the synthetic data is extended to several extra spaces to match with the real data, which leads to a considerable performance improvement.
By adding the Dual Domain Training module (Base+A+T), the information contained in the frequency domain is introduced into the framework, which leads to a big performance improvement.
The improvement is significant in the dataset that is easy to classify in the frequency domain such as Insect and FD.
By adding the Dual Objectives Matching module (Base+A+T+M), the matching between synthetic data and real data is strengthened by the parameter space matching loss, which leads to a decent performance improvement.
In summary, each module proposed in this paper is effective and contributes to the final performance of CondTSC.


\begin{table}[!t]
\centering
\caption{\bl{Accuracy(\%) of CondTSC on the HAR dataset with $spc=5$ initialized by different methods. Initial means the performance of the initial dataset and Final means the performance of dataset optimized by CondTSC.}}
\label{tab:diffinit}
\resizebox{\linewidth}{!}{
\begin{tabular}{c|c|c|c|c|c|c|c}
\toprule
Methods & Random & \multicolumn{2}{c|}{K-means} & \multicolumn{2}{c|}{Kernel K-means} & \multicolumn{2}{c}{Agglomerative Clustering}\\
\midrule
Metric & - & Cosine & Euclidean & Linear & RBF & Ward & Complete\\
\midrule
\midrule
Initial& 68.73$\pm$1.54 & 73.56$\pm$2.46  & 72.19$\pm$1.83& 73.62$\pm$1.81 & 72.43$\pm$3.09 & 64.47$\pm$1.83 & 60.61$\pm$1.66 \\
Final& 79.89$\pm$2.94& 82.43$\pm$1.86  &82.47$\pm$3.39  &81.90$\pm$1.95  &81.74$\pm$2.77  &80.22$\pm$2.74  &77.13$\pm$2.84  \\
Improvement&16.23\% & 12.05\% & 14.24\% & 11.24\% & 12.85\% & 24.42\% & 25.60\% \\
\bottomrule
\end{tabular}
}
\vspace{-.3cm}
\end{table}

\subsection{Case study}

In this section, we will investigate why CondTSC outperforms other baselines in the time series dataset condensation task.
We randomly select some data samples of the Insect dataset as the initial condensed synthetic dataset $\mathcal{S}$ and use CondTSC and MTT (strongest baseline) to train $\mathcal{S}$ separately.
Then, we visualize the training process of a random sample in $\mathcal{S}$ in both the frequency domain and the time domain, which is shown in Fig.~\ref{fig:1visualize}.

From Fig.~\ref{fig:1visualize}, we could observe that:
(1) In the frequency domain, the synthetic data trained by CondTSC exhibits a tendency to conform to the distribution of the original data (as shown in the dashed rectangular boxes in the figure).
The synthetic data (red solid line) are to fit the mean of the real data (black dashed line) and conform to the standard deviation (green vertical segment) simultaneously.
This conformity is subsequently manifested in the time domain.
We can see significant changes in the data of the time domain since signals of various frequencies are modified.
(2) However, the synthetic data trained by MTT does not vary much, especially in the frequency domain.
Moreover, we could observe that the initial high-frequency signals, which deviate from the underlying distribution of the original data, are not modified and incorporated into the final synthetic data.
(3) This significant difference in the learned dataset leads to different results, while CondTSC achieves remarkable performance.


\subsection{Different Initialization}

In this section, we would like to investigate the influence of different initializations of CondTSC.
We try different initializations of different metrics including random selection, K-means, Kernel K-means, and Agglomerative Clustering.
We show the performance of the initial condensed dataset and the final condensed dataset of 5 runs in Table~\ref{tab:diffinit}, from which the following observations could be made.
(1) The performance of different initialized condensed datasets is improved.
This indicates that CondTSC is effective and could improve the performance of various initializations. 
(2) CondTSC utilizes the K-means method, which gets relatively good performance.
(3) Even with the random selection initialization, CondTSC still produces a comparable condensed dataset.
This indicates that CondTSC is an initialization-independent framework.

\section{Conclusion}
In this work, we propose a novel framework that incorporates the time domain and the frequency domain for the time series dataset condensation task.
The proposed framework CondTSC has three novel modules, and each module is designed with different purposes and contributes to the final performance.
Extensive experiments verify the effectiveness of our proposed framework.
In our future work, we will aim to further analyze condensed data and improve the performance of CondTSC.

\section*{Acknowledgment}
We wish to express our sincere appreciation to the researchers of~\cite{UCRArchive2018} for their exceptional efforts in creating, curating, and maintaining this invaluable collection of time series datasets. This work was sponsored by National Key Research and Development Program of China under Grant No.2022YFB3904204, National Natural Science Foundation of China under Grant No. 62102246, 62272301, 62176243, and Provincial Key Research and Development Program of Zhejiang under Grant No. 2021C01034. Part of the work was done when the students were doing internships at Yunqi Academy of Engineering. 
\bibliographystyle{ACM-Reference-Format}
\balance
\bibliography{ref}
\appendix

\section{Surrogate Objective Matching}

\begin{table*}[!th]
\centering
\caption{Cross-Architecture \bl{accuracy(\%)} performance. We train the synthetic data in one network structure and evaluate the synthetic data in another network structure. The experiment is conducted in the HAR dataset with $spc=5$. DD is not selected as one of the baselines due to its bad performance. The best result is highlighted in bold and grey in each column.}
\label{tab:cross}
\resizebox{0.90\textwidth}{!}{
\begin{tabular}{c |c | c | c | c | c | c | c | c | c| c}
\toprule
Train Model& Methods& MLP & CNNBN & CNNIN & ResNet18 & TCN & LSTM & Transformer & GRU \\
\midrule
\midrule

\multirow{7}{*}{MLP}
 & DC & \cellcolor{lgray}{\textbf{60.64$\pm$4.46}} & 67.98$\pm$3.66 & 55.08$\pm$4.65 & 31.35$\pm$8.44 & 59.85$\pm$7.64 & 47.50$\pm$7.92 & 72.12$\pm$4.26& 55.85$\pm$7.05\\
 & DSA &  57.60$\pm$5.98& 67.89$\pm$3.35&53.02$\pm$2.56 & 30.89$\pm$5.24& 67.23$\pm$4.03&45.59$\pm$12.5 & 73.55$\pm$3.62& \cellcolor{lgray}{\textbf{57.01$\pm$7.42}}\\
 & DM & 48.79$\pm$0.97 & 63.75$\pm$3.01 & 59.16$\pm$1.01 & 51.68$\pm$2.70 & 61.39$\pm$2.89 & \cellcolor{lgray}{\textbf{48.60$\pm$0.95}}& 72.04$\pm$2.71 & 51.43$\pm$2.81\\
 & MTT & 51.17$\pm$0.79 &  63.36$\pm$1.29& 58.51$\pm$1.47 & 59.66$\pm$2.43 & 50.97$\pm$3.76&43.50$\pm$5.42 &69.67$\pm$2.52 & 49.10$\pm$1.44\\
 & IDC & 47.67$\pm$1.75 & 54.61$\pm$7.40 & 49.76$\pm$8.28 & 50.93$\pm$7.32 &49.67$\pm$3.32 & 42.44$\pm$2.12 &  65.88$\pm$2.68 & 50.40$\pm$1.09\\
 & HaBa & 40.40$\pm$6.31 & 41.35$\pm$9.39 & 40.25$\pm$5.13  & 34.06$\pm$9.57 & 39.52$\pm$7.12 & 35.42$\pm$3.11 & 49.41$\pm$5.62 & 25.74$\pm$4.53 \\
  \cline{2-10}
 & CondTSC & 52.53$\pm$3.71 & \cellcolor{lgray}{\textbf{73.95$\pm$1.96}} & \cellcolor{lgray}{\textbf{65.40$\pm$3.43}} & \cellcolor{lgray}{\textbf{75.55$\pm$1.75}} & \cellcolor{lgray}{\textbf{67.49$\pm$1.58}} & 45.90$\pm$2.04 & \cellcolor{lgray}{\textbf{73.97$\pm$1.10}} & 51.63$\pm$1.26\\
\midrule
\multirow{7}{*}{CNNBN} 
 & DC & 50.60$\pm$1.92 & 65.83$\pm$3.97 & 42.07$\pm$6.85 & 22.74$\pm$7.06 & 70.89$\pm$5.25& 45.39$\pm$9.78 & 73.39$\pm$4.24 & 54.18$\pm$3.85\\
 & DSA & 48.81$\pm$1.27 &65.16$\pm$3.67 &52.03$\pm$5.96 &29.40$\pm$7.94 & 71.76$\pm$4.54&50.43$\pm$8.53 &75.84$\pm$3.93 &59.97$\pm$3.45 \\
 & DM &  49.21$\pm$1.02& 70.82$\pm$4.63 & 65.57$\pm$3.09 & 64.92$\pm$2.17 & 65.68$\pm$3.34& 45.91$\pm$6.34 &74.57$\pm$3.03 & 56.13$\pm$7.49\\
 & MTT & 50.38$\pm$1.16 &77.66$\pm$3.86 & 58.95$\pm$1.86 & 68.73$\pm$4.42 & 67.86$\pm$2.84 & 49.71$\pm$7.36 & 77.55$\pm$1.58 & 56.35$\pm$3.92\\
 & IDC & 47.58$\pm$1.93 & 54.41$\pm$5.02 & 45.02$\pm$5.67 & 51.03$\pm$4.42 & 45.36$\pm$5.80 &  40.60$\pm$3.50& 56.83$\pm$7.86 & 40.60$\pm$3.50 \\
 & HaBa & 44.90$\pm$1.50 & 56.22$\pm$5.56 & 50.25$\pm$2.27 & 54.21$\pm$7.04 & 53.84$\pm$2.56 & 39.98$\pm$5.58 & 61.56$\pm$6.56 & 47.60$\pm$3.63 \\
  \cline{2-10}
 & CondTSC & \cellcolor{lgray}{\textbf{51.92$\pm$3.36}} & \cellcolor{lgray}{\textbf{82.20$\pm$2.68}}  & \cellcolor{lgray}{\textbf{72.33$\pm$2.93}} &\cellcolor{lgray}{\textbf{76.02$\pm$3.48}} & \cellcolor{lgray}{\textbf{76.95$\pm$2.77}} &\cellcolor{lgray}{\textbf{53.16$\pm$4.32}} & \cellcolor{lgray}{\textbf{79.81$\pm$1.93}}  &\cellcolor{lgray}{\textbf{60.37$\pm$2.61}}  \\
\midrule
\multirow{7}{*}{CNNIN}
 & DC & 45.41$\pm$4.45 & 57.64$\pm$6.92& 52.84$\pm$2.71 & 25.93$\pm$7.38 & 66.85$\pm$3.36 & 36.40$\pm$5.84 & 46.53$\pm$5.03 & 35.50$\pm$1.79\\
 & DSA & 51.27$\pm$2.19 &61.01$\pm$4.43 &53.08$\pm$2.56 &27.43$\pm$8.02 & 66.94$\pm$3.29&48.78$\pm$10.1 & 74.58$\pm$2.86& \cellcolor{lgray}{\textbf{55.33$\pm$5.00}}\\
 & DM &  47.37$\pm$2.30 & 57.87$\pm$2.33 & 57.13$\pm$4.08 &58.24$\pm$2.67 &56.77$\pm$3.19 &43.72$\pm$8.05 &57.72$\pm$2.70 &39.81$\pm$4.24 \\
 & MTT & 51.22$\pm$1.45 & 64.61$\pm$1.94 & 58.20$\pm$2.10 & 66.13$\pm$2.56 & 64.30$\pm$1.57 & 46.94$\pm$9.39 & 73.59$\pm$1.84 & 48.84$\pm$4.02 \\
 & IDC & 43.20$\pm$0.93 & 52.57$\pm$6.55 & 48.00$\pm$4.09 & 52.13$\pm$5.33 & 44.56$\pm$6.41 & 45.47$\pm$4.93 & 55.67$\pm$3.62 & 41.75$\pm$2.84 \\
 & HaBa & 49.54$\pm$1.30 & 27.35$\pm$6.46 & 59.44$\pm$2.74 & 29.41$\pm$7.66 & 46.89$\pm$6.47 & 37.12$\pm$2.19 &  53.16$\pm$7.89 & 41.47$\pm$4.09\\
  \cline{2-10}
 & CondTSC & \cellcolor{lgray}{\textbf{54.52$\pm$3.33}}  & \cellcolor{lgray}{\textbf{74.37$\pm$2.56}}  & \cellcolor{lgray}{\textbf{71.60$\pm$2.58}} & \cellcolor{lgray}{\textbf{70.42$\pm$3.94}}  & \cellcolor{lgray}{\textbf{67.09$\pm$1.38}}  & \cellcolor{lgray}{\textbf{49.12$\pm$4.29}}  & \cellcolor{lgray}{\textbf{75.64$\pm$2.10}} &  49.23$\pm$1.04\\
 \midrule
 \multirow{7}{*}{ResNet18}
 & DC & 17.22$\pm$2.21 & 59.26$\pm$4.03& 52.78$\pm$3.64 & 30.07$\pm$6.06 & 64.32$\pm$1.94 & 21.68$\pm$7.34 & 69.90$\pm$6.81 & 18.25$\pm$3.10\\
 & DSA &  46.72$\pm$4.31& 59.91$\pm$4.27& 51.93$\pm$2.99&32.36$\pm$8.29 &66.25$\pm$5.69 & \cellcolor{lgray}{\textbf{47.65$\pm$10.5}}&70.95$\pm$5.27 & 38.33$\pm$6.82\\
 & DM & 35.80$\pm$1.94 &49.01$\pm$2.86 & 49.95$\pm$2.11&50.16$\pm$2.61 &43.28$\pm$4.27 & 39.32$\pm$9.09&34.01$\pm$7.36 & 43.64$\pm$4.00\\
 & MTT & 49.75$\pm$0.60 &  67.42$\pm$1.55 & 57.87$\pm$1.90 & 65.43$\pm$2.50 & 50.82$\pm$3.62 & 43.20$\pm$2.02 & 73.50$\pm$2.21 &  44.28$\pm$3.33\\
 & IDC & 45.11$\pm$1.16 & 44.27$\pm$5.39 & 28.35$\pm$6.03 & 42.35$\pm$6.28 & 38.59$\pm$4.08 & 42.79$\pm$7.10 & 49.46$\pm$4.44 & 42.45$\pm$6.20 \\
 & HaBa & 31.04$\pm$4.93 & 19.92$\pm$5.23 & 33.59$\pm$6.04 & 18.09$\pm$1.45 & 30.25$\pm$6.32 & 35.00$\pm$9.14 & 29.77$\pm$8.47 &  30.66$\pm$5.86\\
  \cline{2-10}
 & CondTSC & \cellcolor{lgray}{\textbf{53.68$\pm$2.02}}  & \cellcolor{lgray}{\textbf{76.53$\pm$1.46}}  &\cellcolor{lgray}{\textbf{70.08$\pm$3.74}}  & \cellcolor{lgray}{\textbf{77.67$\pm$2.09}}  & \cellcolor{lgray}{\textbf{69.85$\pm$0.95}} & 44.56$\pm$3.79  & \cellcolor{lgray}{\textbf{77.24$\pm$2.38}}  & \cellcolor{lgray}{\textbf{45.02$\pm$1.69}}  \\
 \midrule
\multirow{7}{*}{TCN}
 & DC & 47.75$\pm$1.93 & 59.42$\pm$3.01 & 53.97$\pm$3.38 & 31.07$\pm$7.48& 64.02$\pm$4.58 &48.85$\pm$11.7 &71.76$\pm$3.12 & 54.04$\pm$5.62 \\
 & DSA &  47.71$\pm$1.63& 60.92$\pm$3.42& 55.30$\pm$3.17&30.14$\pm$6.29 & 64.31$\pm$4.00& 45.03$\pm$13.2& 70.15$\pm$5.12& \cellcolor{lgray}{\textbf{54.11$\pm$3.40}}\\
 & DM & 48.21$\pm$1.52 & 59.66$\pm$6.08& 57.17$\pm$1.95&50.98$\pm$2.61 &63.67$\pm$2.66 & 48.59$\pm$10.3  & 70.08$\pm$5.21& 51.48$\pm$3.44\\
 & MTT & 50.59$\pm$0.68 & 65.70$\pm$1.50 & 59.25$\pm$1.50 & 61.65$\pm$2.25 & 58.33$\pm$2.82 & 47.29$\pm$5.12 & 70.48$\pm$2.01 & 50.46$\pm$2.88\\
 & IDC & 48.29$\pm$1.77 & 58.92$\pm$3.92 & 51.82$\pm$1.91 & 57.12$\pm$7.36 & 49.91$\pm$2.47 & 47.59$\pm$1.40 & 67.11$\pm$4.49 & 48.42$\pm$1.48\\
 & HaBa & 36.76$\pm$2.41  & 34.68$\pm$5.41 & 35.61$\pm$5.12 & 35.91$\pm$7.07 & 25.53$\pm$5.54 & 29.85$\pm$3.03 &  39.54$\pm$7.14 & 27.20$\pm$4.95\\
 \cline{2-10}
 & CondTSC & \cellcolor{lgray}{\textbf{52.64$\pm$2.95}} & \cellcolor{lgray}{\textbf{71.01$\pm$3.54}}  & \cellcolor{lgray}{\textbf{66.45$\pm$2.49}}  & \cellcolor{lgray}{\textbf{67.96$\pm$4.04}}  & \cellcolor{lgray}{\textbf{65.67$\pm$1.74}}  & \cellcolor{lgray}{\textbf{48.88$\pm$2.63}} & \cellcolor{lgray}{\textbf{71.93$\pm$3.47}}  &51.29$\pm$1.34 \\
\midrule

\bottomrule
\end{tabular}
}
\end{table*}

In this section, we introduce the details of objective matching.
To learn the condensed synthetic dataset $\mathcal{S}$, we would match some surrogate objectives of $\mathcal{S}$ and $\mathcal{\Tau}$. 
The aim is to ensure that the synthetic dataset $\mathcal{S}$ accurately captures the dynamics present in the real dataset $\mathcal{\Tau}$.
Consequently, a network trained using the synthetic dataset can exhibit comparable performance to a network trained using the complete real dataset.
Here, we introduce two types of surrogate objectives used in this paper.
\subsection{Gradient Matching}

The training gradient for a given initial model trained by $\mathcal{S}$ and $\mathcal{\Tau}$ is a good surrogate objective.
By matching the training gradients of $\mathcal{S}$ and $\mathcal{\Tau}$, we can effectively align the training dynamics of the synthetic dataset with that of the real dataset~\cite{zhao2020dataset,zhao2021dataset,du2023minimizing}. 
The single-step gradient matching objective is shown as follows.
\begin{equation}
    S^*=\mathop{\arg\min}\limits_\mathcal{S}\mathbb{E}_{{\theta_0}\sim p(\theta)}[\textbf{D}(\nabla_{\theta_0}\ell(f_{\theta_0}(\mathcal{S})), \nabla_{\theta_0}\ell(f_{\theta_0}(\mathcal{\Tau})))]
    \label{eq:GM}
\end{equation}
Here $\textbf{D}(\cdot,\cdot)$ is the distance function such as cosine distance and $p(\theta)$ is the distribution of the network parameter.
$\nabla_{\theta_0}\ell(f_{\theta_0}(\mathcal{S}))$ is the gradient of the training loss $\ell(f_{\theta_0}(\mathcal{S}))$ of network $f$ parameterized by $\theta_0$ on dataset $\mathcal{S}$.
This equation aims to find the synthetic data $\mathcal{S}$ that has the most similar gradient to real data $\mathcal{\Tau}$ with a given initial network $f_{\theta_0}$.
In practice, the synthetic data $\mathcal{S}$ would be updated by many networks $f_{\theta_0}$ to ensure the ability of generalization.

\subsection{Distribution Matching}
Another surrogate objective is the data distribution generated by a given model based on $\mathcal{S}$ and $\mathcal{\Tau}$.
Its essential aim is that the network trained by $\mathcal{S}$ could generate similar embeddings to those from the network trained by $\mathcal{\Tau}$.
Consequently, the data distributions of $\mathcal{S}$ and $\mathcal{\Tau}$ are similar, which makes $\mathcal{S}$ a condensed dataset from $\mathcal{\Tau}$.
The core idea of distribution matching is reducing the empirical estimate of the Maximum Mean Discrepancy~\cite{zhao2023dataset}, which is shown as follows.
\begin{equation}
    S^*=\mathop{\arg\min}\limits_\mathcal{S}\mathbb{E}_{{\theta_0}\sim p(\theta)}[||\cfrac{1}{\mathcal{|S|}}\sum_{i=1}^\mathcal{|S|}f_{\theta_0}(s_i) - \cfrac{1}{|\mathcal{\Tau}|}\sum_{j=1}^{|\mathcal{\Tau}|}f_{\theta_0}(x_j)||^2]
    \label{eq:DM}
\end{equation}
Here $p(\theta)$ is the distribution of the network parameter and $f_{\theta_0}(\cdot)$ is the network $f$ parameterized by $\theta_0$ which outputs the embedding of the input.
In practice, the calculation of Eq.~\ref{eq:DM} is usually between a batch of $\mathcal{S}$ and a batch of $\mathcal{\Tau}$.
This equation is essentially finding the synthetic data $\mathcal{S}^*$ that has the most similar distribution to the full data $\mathcal{\Tau}$ in the network embedding space $f_{\theta}$.

By matching the surrogate objectives, the models trained by the condensed synthetic data could be similar to the models trained by the full real data.
Consequently, training a model with the condensed synthetic data results in minimal performance degradation and is much more efficient simultaneously.

\section{Experiment details and efficiency}
\label{sec:exp_details}

\noindent{\textbf{Network Structure:}}
In line with prior research~\cite{nguyen2021dataset,wang2020dataset,liu2022dataset,deng2022remember,zhou2022dataset,loo2023dataset,jin2021graph,jin2022condensing,kim2022dataset,zhao2021dataset,zhao2020dataset,zhao2023dataset} that utilize simple deep neural networks to verify the effectiveness of the proposed method, we train the synthetic data based on a wide range of basic network architectures, including multi-layer perception (MLP), convolution neural network (CNN), ResNet~\cite{he2016deep}, temporal convolutional network (TCN)~\cite{bai2018empirical}, LSTM~\cite{hochreiter1997long}, Transformer~\cite{vaswani2017attention}, gate recurrent unit (GRU)~\cite{chung2014empirical}. 
For MLP, we use a 3-layer MLP with a hidden size of 128 and a ReLU~\cite{nair2010rectified} activation function.
For CNN, we use a 3-layer CNN with hidden channels of 128, kernel size of 3x1, maxpooling function, and ReLU activation function.
Specifically, following the previous studies~\cite{zhou2022dataset}, we denote CNN with batch normalization~\cite{ioffe2015batch} and instance normalization~\cite{ulyanov2017instance} as CNNBN and CNNIN respectively.
For ResNet, we use the default network structure of Resnet18.
For TCN, we use a 1-layer TCN with a kernel size of 64x1.
For LSTM, we use a 1-layer LSTM with a hidden size of 100.
For Transformer, we use a 1-layer Transformer with a hidden size of 64.
For GRU, we use a 1-layer GRU with a hidden size of 100.
These networks except ResNet have a similar size of model parameters, which is approximately 30,000.

\noindent\textbf{Hyper-parameter Searching of Baselines:}
We use the following hyper-parameter searching strategy to fairly evaluate baseline methods. It is hard to explore the same grid search space due to the time limit (one baseline might take dozens of days to search). Consequently, for each method, we first search for a good range for each parameter in the same log space while fixing other parameters. Then we use the searched range to do a grid search as follows.
we choose the epoch with the highest validation accuracy for testing.
Consequently, we believe the comparisons are fair.

\begin{figure}[!th]
    \centering
    \includegraphics[width=0.95\linewidth]{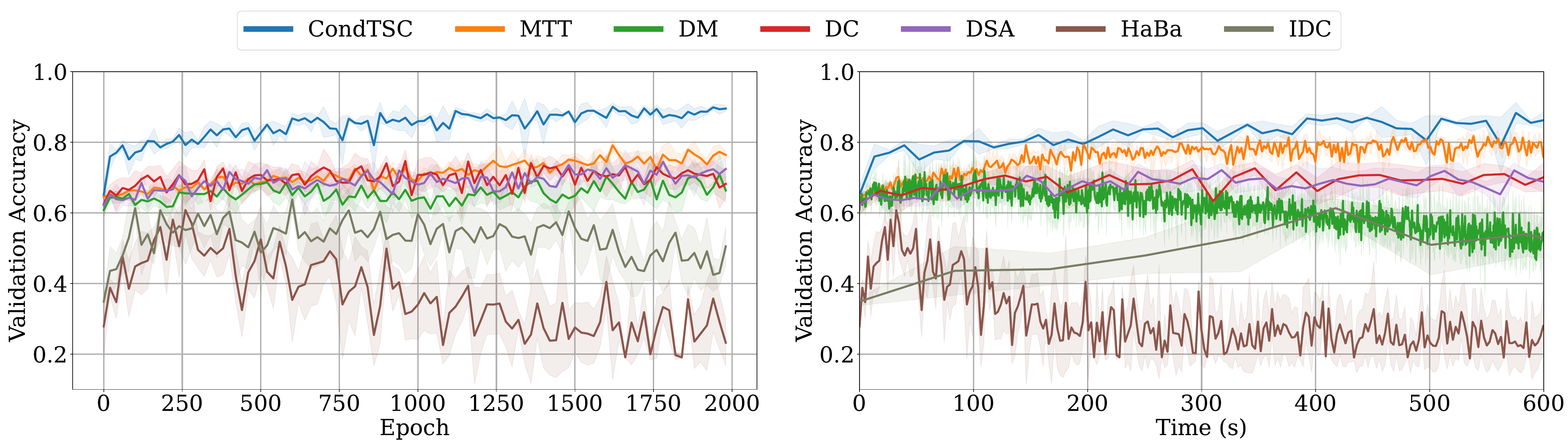}
    \caption{The training curve on the HAR dataset with $spc=5$. We show the training epoch and training time as the x-axis respectively. The mean and standard deviation of validation accuracy of 5 runs are shown.}
    \label{fig:6_together}
\end{figure}

\begin{itemize}[leftmargin=*]
    \item \textbf{DD:} Training epochs is 100 and evaluate every 2 epochs. The learning rates of the synthetic data are searched in \bl{[$10^{-3}$, $10^{-2}$, $10^{-1}$, 1, 10]}. The distilled steps are searched in [1, 5, 10, 20, 30, 40], and the distilled epochs are searched in [1, 2, 3, 4, 5].
    \item \textbf{DC, DSA, DM:} Training epochs is 2000 and evaluate every 100 epochs. The feature learning rate are searched in \bl{[$10^{-3}$, $10^{-2}$, $10^{-1}$, 1, 10]}, and the net learning rate is searched in \bl{[$10^{-3}$, $10^{-2}$, $10^{-1}$, 1, 10, 100, 1000]}.
    \item \textbf{MTT:} Training epochs is 2000 and evaluate every 100 epochs. The learning rate are searched in \bl{[$10^{-3}$, $10^{-2}$, $10^{-1}$, 1, 10, 100, 1000]}. The $\alpha$ learning rate is searched in \bl{[$10^{-9}$, $10^{-8}, 10^{-7}$, $10^{-6}$]}.
    \item \textbf{IDC:} Training epochs is 100 and evaluate every 2 epochs. The matching surrogate objective is set to feature matching. The multi-formation factor is searched in [1, 2, 3], and the learning rate is searched in \bl{[$10^{-3}$, $10^{-2}$, $10^{-1}$, 1, 10, 100, 1000]}.
    \item \textbf{HaBa:} Training epochs is 2000 and evaluate every 100 epochs. The learning rates of features and styles are searched in \bl{[$10^{-3}$, $10^{-2}$, $10^{-1}$, 1, 10, 100, 1000]}. The learning rate of $\alpha$ is searched in \bl{[$10^{-9}$, $10^{-8}$, $10^{-7}$, $10^{-6}$]}.
\end{itemize}

All of the hyper-parameters that are not mentioned above are kept the same as in the original paper.
Moreover, we show the training curve of different methods with respect to training epochs and time in Fig.~\ref{fig:6_together}.
The experiment is conducted on the HAR dataset with $spc=5$ and the mean and standard deviation of validation accuracy of 5 runs are shown.
We can observe that it only costs 10 minutes to train CondTSC to converge.
Moreover, We can observe that CondTSC achieves the best training accuracy no matter with the same training epoch or the same training time.
Furthermore, the training curve indicates all of the baseline methods are converged and we choose the epoch with the highest validation accuracy for testing.
Consequently, we believe the comparisons are fair.

\section{Cross Architecture Performance}

In the dataset condensation task, the ability of cross-architecture generalization is one of the important indicators for evaluating methods~\cite{wang2020dataset,zhao2020dataset}.
This is because when training the synthetic data $\mathcal{S}$, we usually utilize only one network architecture~\cite{cazenavette2022dataset,kim2022dataset,lee2022dataseta,lee2022datasetb,zhao2021dataset}, i.e., $p(\theta)$ is from a parameter space of single network structure $f$.
Consequently, the learned synthetic data performs well when training the network $f$ but might perform satisfactorily when training the network with another structure $g$.
To verify the ability to cross-architecture generalization of the learned synthetic data of CondTSC, we conduct extensive experiments on the HAR dataset with $spc=5$.
We train the synthetic data $\mathcal{S}$ with one network structure and evaluate $\mathcal{S}$ with another network structure.
The result is shown in Table~\ref{tab:cross}.

From Table~\ref{tab:cross}, we could observe that:
(1) Overall, CondTSC performs best in the cross-architecture experiment.
It achieves stable and outstanding performance, especially in the CNN-based and Transformer-based models.
This indicates that CondTSC learns a robust and easy-to-generalize synthetic dataset and is resistant to overfitting one network architecture.
(2) In some simple network architectures such as MLP and RNN, the single-step gradient matching methods and distribution matching methods, i.e., DC, DSA, and DM perform relatively well due to their simple design.
However, their performance is unstable, which indicates that they could not learn a robust synthetic dataset and lack the ability to generalize to other network architectures.

\section{Parameter Sensitivity}

\begin{figure*}[!t]
    \centering
    \includegraphics[width=1\linewidth]{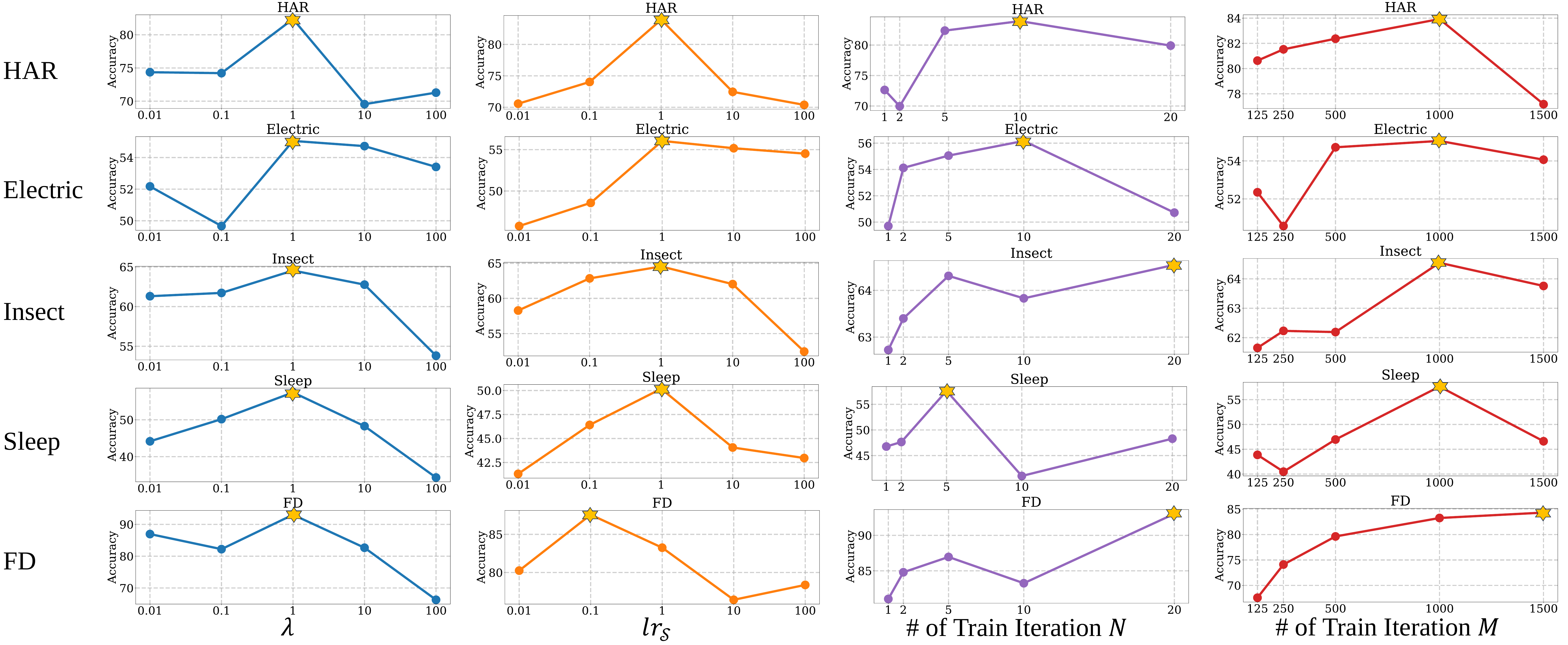}
    \caption{The parameter sensitivity analysis. 
    Different values of hyper-parameters are evaluated on all five datasets. 
    We evaluate four hyper-parameters, including loss scaler $\lambda$, the learning rate of the synthetic data $lr_{\mathcal{S}}$, the number of train iterations for the synthetic data $N$, the number of train iterations for the real data $M$.
    The yellow star indicates the hyper-parameter value with the highest test accuracy.}
    \label{fig:4param}
\end{figure*}

In this section, we would like to show the performance of CondTSC on different hyper-parameters, which throws light on the analysis of CondTSC.
We evaluate the performance of different values of hyper-parameters on all five datasets, and the results are shown in Fig.~\ref{fig:4param}.
(1) We evaluate the performance of different loss scaler $\lambda$, which balances the gradient matching loss $\mathcal{L}^{grad}$ and the embedding matching loss $\mathcal{L}^{emb}$.
We test a range of log-scale values and observe setting $\lambda=1$ is best for all datasets.
(2) We evaluate the performance of different learning rates for the synthetic data $lr_{\mathcal{S}}$.
We also test a range of log-scale values for $lr_{\mathcal{S}}$.
In previous studies on image dataset condensation~\cite{zhao2021dataset,zhao2023dataset,cazenavette2022dataset,liu2022dataset}, the value of $lr_{\mathcal{S}}$ are usually set to a large value, i.e., $lr_{\mathcal{S}}\geq1e3$.
However, we could observe that setting $lr_{\mathcal{S}}=1$ is best for our time series dataset condensation task.
This demonstrates that the time series dataset condensation task has essential differences from other dataset condensation tasks, and designing special techniques for time series dataset condensation is necessary.
(3) We also evaluate the performance of different values of $N$ and $M$, which is the number of training iterations in the Dual Domain Training module for the synthetic data $\mathcal{S}$ and the real data $\mathcal{\Tau}$ respectively.
From our analysis in the Method section, we should set $N\ll M$ to avoid overfitting to the real data.
Here, the experiment results show that the performance is relatively good when $N\approx10$ and $M\approx1000$ and the performance could be unsatisfactory when $M\approx N$.
This further validates our analysis.

Overall, we conducted extensive experiments on evaluating the performance of different values of hyper-parameters.
We find that setting the loss scaler $\lambda=1$, the learning rate of synthetic data $lr_{\mathcal{S}}=1$, and training iterations $N\ll M$ is a good hyper-parameter setting.

\end{document}